\documentclass[10pt]{article}

\usepackage{arxiv}
\usepackage{float}
\usepackage{pgfplots}
\usepackage{pgfplotstable}
\pgfplotsset{compat=1.18}
\usetikzlibrary{arrows.meta}

\usetikzlibrary{arrows.meta,positioning,calc,fit,backgrounds}
\pgfdeclarelayer{nodelayer}
\pgfdeclarelayer{edgelayer}
\pgfsetlayers{background,main,nodelayer,edgelayer}

\definecolor{cssniper}{RGB}{34,178,76}
\definecolor{csraw}{RGB}{150,156,162}
\definecolor{csmark}{RGB}{224,91,91}
\definecolor{csband}{RGB}{205,228,250}
\definecolor{cswaste}{RGB}{252,226,226}
\definecolor{csprompt}{RGB}{242,244,247}
\definecolor{cssufficient}{RGB}{220,242,226}
\definecolor{csnoise}{RGB}{224,91,91}
\definecolor{csink}{RGB}{34,39,47}
\definecolor{csblue}{RGB}{80,128,208}
\definecolor{csviolet}{RGB}{138,99,210}
\definecolor{cswarn}{RGB}{230,150,40}

\pgfplotsset{
  csfig/.style={
    tick label style={font=\footnotesize\rmfamily},
    label style={font=\footnotesize\rmfamily},
    legend style={
      font=\footnotesize\rmfamily,
      draw=none,
      fill=white,
      fill opacity=0.92,
      cells={anchor=west},
    },
  },
  csbar/.style={ybar=1.5pt, bar width=8pt},
  csline-snipe/.style={csblue, line width=1.8pt, smooth},
  csline-raw/.style={csraw, very thick, smooth},
}

\tikzset{
  cs/base/.style={
    rounded corners=4pt,
    line width=0.45pt,
    font=\scriptsize\rmfamily,
    align=center,
    inner sep=4pt,
  },
  cs/panel/.style={
    cs/base,
    rounded corners=6pt,
    draw=csraw!35,
    fill=csprompt,
    inner sep=6pt,
  },
  cs/title/.style={font=\footnotesize\bfseries\rmfamily, text=csink},
  cs/subtitle/.style={font=\scriptsize\rmfamily, text=csink!70},
  cs/micro/.style={font=\tiny\rmfamily, text=csraw!70!black},
  cs/tool/.style={cs/base, draw=csraw!55, fill=white, minimum height=0.34cm},
  cs/tool-raw/.style={cs/tool, draw=csraw!70, fill=csraw!10},
  cs/tool-good/.style={cs/tool, draw=csblue!65, fill=csblue!8},
  cs/process/.style={cs/base, draw=csraw!55, fill=white, minimum height=0.55cm},
  cs/process-good/.style={cs/process, draw=csblue!65, fill=csblue!7},
  cs/process-warn/.style={cs/process, draw=csnoise!65, fill=cswaste!75},
  cs/evidence/.style={cs/base, draw=cssniper!70, fill=cssniper!8},
  cs/noise/.style={cs/base, draw=csnoise!70, fill=csnoise!8},
  cs/arrow/.style={
    -{Stealth[length=\csWfArrowTip, width=\csWfArrowWidth]},
    line width=\csWfArrowLW,
    draw=csraw!75,
    shorten <=\csWfArrowShorten,
    shorten >=\csWfArrowShorten,
  },
  cs/good-arrow/.style={
    -{Stealth[length=\csWfArrowTip, width=\csWfArrowWidth]},
    line width=\csWfArrowLW,
    draw=csblue!80!black,
    shorten <=\csWfArrowShorten,
    shorten >=\csWfArrowShorten,
  },
  cs/retry-arrow/.style={
    -{Stealth[length=\csWfArrowTip, width=\csWfArrowWidth]},
    line width=\csWfArrowLW,
    draw=csnoise!80!black,
    dashed,
    shorten <=\csWfArrowShorten,
    shorten >=\csWfArrowShorten,
  },
}

\tikzset{
  wf panel/.style={
    rounded corners=6pt,
    draw=csraw!35,
    fill=csprompt,
    inner sep=6pt,
    minimum width=7.6cm,
    minimum height=5.70cm,
    anchor=south west,
    font=\scriptsize\rmfamily,
  },
  wf process/.style={
    rounded corners=4pt,
    draw=csraw!55,
    fill=white,
    line width=0.45pt,
    font=\scriptsize\rmfamily,
    align=center,
    inner sep=4pt,
    minimum height=0.55cm,
  },
  wf process system/.style={wf process, draw=csblue!65, fill=csblue!7},
  wf process warn/.style={wf process, draw=csnoise!65, fill=cswaste!75},
  wf tool raw/.style={
    rounded corners=4pt,
    draw=csraw!70,
    fill=csraw!10,
    line width=0.45pt,
    font=\scriptsize\rmfamily,
    align=center,
    inner sep=4pt,
    minimum height=0.34cm,
  },
  wf tool system/.style={wf tool raw, draw=csblue!65, fill=csblue!8},
  wf label/.style={font=\scriptsize\rmfamily, text=csink!70, inner sep=0pt, draw=none, fill=none},
  wf micro/.style={font=\tiny\rmfamily, text=csraw!70!black, inner sep=0pt, draw=none, fill=none},
  wf arrow/.style={
    -{Stealth[length=1.4mm, width=1.05mm]},
    line width=0.45pt,
    draw=csraw!75,
    shorten <=1.4pt,
    shorten >=1.4pt,
  },
  wf system arrow/.style={
    -{Stealth[length=1.4mm, width=1.05mm]},
    line width=0.45pt,
    draw=csblue!80!black,
    shorten <=1.4pt,
    shorten >=1.4pt,
  },
}

\newcommand{\csLegendSwatch}[2]{%
  \tikz[baseline=-0.5ex]\draw[draw=#1!70!black, fill=#1!35, rounded corners=1pt] (0,0) rectangle (0.22,0.10);%
  \hspace{0.25em}{\scriptsize\rmfamily #2}%
}

\newcommand{\csMiniLegend}{%
  {\csLegendSwatch{cssniper}{useful evidence}\hspace{0.9em}%
   \csLegendSwatch{csraw}{neutral context}\hspace{0.9em}%
   \csLegendSwatch{csnoise}{noise}}%
}

\newcommand{\csStripSeg}[5]{%
  \draw[draw=#4!70!black, fill=#4!35, rounded corners=1pt] (#1,#2) rectangle ++(#3,\csWfStripH);
  \if\relax\detokenize{#5}\relax\else
    \node[cs/micro, text=#4!55!black, inner sep=0pt] at ($(#1,#2)+(#3/2,\csWfStripH/2)$) {#5};
  \fi
}

\def\csWfArrowTip{1.4mm}
\def\csWfArrowWidth{1.05mm}
\def\csWfArrowLW{0.45pt}
\def\csWfArrowShorten{1.4pt}

\def\csWfStripH{0.20}

\usepackage[square,comma,numbers,sort&compress]{natbib}
\let\parencite\citep

\let\oldbibliography\thebibliography
\def\thebibliography#1{\oldbibliography{#1}\setlength{\itemsep}{0.15\baselineskip}}

\title{ContextSniper: AntTrail's Token-Efficient Code Memory for Repository-Level Program Repair}

\author{
  \begin{tabular}{c}
    Chiwang~Luk\textsuperscript{1,\textdagger} \quad
    Matin~Mohammad~Najafi\textsuperscript{1} \quad
    Zhifeng~Jia\textsuperscript{1}\\
    Wei~Yang\textsuperscript{1} \quad
    Xiuchang~Li\textsuperscript{1} \quad
    Jinwei~Zhu\textsuperscript{1}\\
    Yang~Ren\textsuperscript{1} \quad
    Lei~Chen\textsuperscript{2} \quad
    Gao~Cong\textsuperscript{3,\textdagger}\\[0.35em]
    {\small \textsuperscript{1}Huawei \quad
    \textsuperscript{2}HKUST(GZ) \quad
    \textsuperscript{3}Nanyang Technological University}
  \end{tabular}
}

\date{}

\begin{document}

\maketitle
\begingroup
\renewcommand{\thefootnote}{\textdagger}
\footnotetext{Corresponding authors.}
\endgroup

\begin{abstract}
Large language model agents can repair real repository issues, but they often spend large context budgets on whole-file reads, broad searches, and long terminal outputs where useful evidence is mixed with irrelevant code and logs. This paper presents ContextSniper, AntTrail's code-repair module for precision evidence selection in repository-level program repair, part of AntTrail's broader agent-memory engine\footnote{AntTrail: \href{https://gitcode.com/datagallery/AntTrail}{gitcode.com/datagallery/AntTrail}.}. ContextSniper indexes code and action memory as three views (compact abstract, structured overview, full content), retrieves candidates with a hybrid ranker that fuses semantic embeddings, BM25, symbolic (ctags-style) metadata, and graph relations, filters long tool output through an intention-aware context gate, and returns compact evidence packets while keeping full source recoverable on demand. In a matched 50-task-per-condition comparison on SWE-bench Lite (same tasks, baseline vs.\ ContextSniper), ContextSniper reduces total token use by 51.5\% and logged cost by 36.4\% for OpenClaw, with a comparable submitted-resolution rate (a difference of one task per 50). In a separate five-task comparison, ContextSniper beats existing memory- and RAG-style integrations (mem0, Letta/MemGPT, Serena, LlamaIndex-style RAG) on token efficiency. These results suggest a dedicated context-access layer can substantially cut token and cost overhead for repository-level repair agents without a measurable loss in repair quality.\footnote{Evaluation harness for this study: \href{https://github.com/Calluking/ContextSniper}{github.com/Calluking/ContextSniper}.}
\end{abstract}


\section{Introduction}
\fontsize{10}{12}\selectfont

\subsection{Motivation}
When a coding agent cannot find the evidence it needs, it does not stop; it rereads. A failed patch attempt on a real repository issue often triggers another whole-file read, another broad search, another look at a stack trace already seen once, producing a read--try--fail--reread loop that spends tokens without necessarily finding the missing evidence \parencite{yang2024sweagent,vogel2026codebasememory}. This failure mode matters because large language model code agents are now evaluated and deployed for exactly this kind of repository-level work: repair benchmarks and agent systems require models to inspect codebases, execute tools, and generate source-code patches rather than only complete isolated snippets \parencite{jimenez2024swebench,yang2024sweagent,zhang2024autocoderover,xia2024agentless}. Production coding agents apply the same inspect-edit-test loop to real development and debugging work. In these workflows, context is not merely auxiliary text: it is the agent's working record of repository state. Losing a single file path, failing-test trace, API precondition, or symbol relationship can derail localization and cause an otherwise plausible patch to target the wrong code, which mirrors recent long-horizon agent work arguing that context must preserve precise actions, observations, and evolving state rather than simply maximize length \parencite{kang2025acon}.

The difficulty is that prevailing agents obtain this evidence through raw repository access. As illustrated in Figure~\ref{fig:contextsniper-overview}A, a task prompt is mixed with whole-file reads, broad search results, and long terminal logs. As interactions accumulate, the active context grows with both useful evidence and stale exploration residue. This growth creates two coupled bottlenecks for repository repair. First, it raises inference cost and latency because every later reasoning step must carry more tokens. Second, it degrades decision quality because the few lines that explain the failure are diluted by unrelated helpers, repeated stack traces, build noise, and earlier failed attempts. Long-context studies show that models may fail to use relevant information reliably when it is buried inside long inputs \parencite{liu2024lostmiddle}; repository repair makes this problem concrete because the evidence needed for a patch is often surrounded by unrelated code and runtime output.

Naive context reduction is not enough. Simple truncation can remove the only test assertion or import edge that explains a bug, while generic summarization can blur provenance, line numbers, and editability. Effective context management for coding agents must reduce aggressively while retaining heterogeneous repair signals: issue intent, code structure, symbol definitions, failing observations, command outcomes, and the updated state after edits. Retrieval-augmented generation and repository-memory systems motivate a separation between knowledge storage and prompt construction \parencite{lewis2020rag,vogel2026codebasememory,ouyang2024repograph}, but repository repair also requires control over what read and command output is allowed to enter the active context in the first place.

This paper builds on a different view of context: evidence, not ballast. AntTrail is a general agent-memory engine for durable task context, designed to store, index, retrieve, and budget the information an agent may need instead of leaving it buried in raw transcripts, whole artifacts, or unfiltered tool output \parencite{anttrail2026gitcode}. ContextSniper is AntTrail's module specialized for repository-level program repair. Figure~\ref{fig:contextsniper-overview}B shows the resulting workflow. Instead of passing raw repository and runtime output directly to the LLM, ContextSniper first treats it as candidate evidence. We describe this evidence-capture process as \emph{sniping}: the system keeps task-relevant code and runtime signals, removes surrounding low-value regions, and assembles the surviving material into a compact evidence packet for the agent.

\pgfplotstableread[col sep=comma]{resource/cross-host-efficiency.csv}\openingcrosshostdata
\pgfplotstableread[col sep=comma]{resource/opening-figure-bars.csv}\openingbardata
\pgfplotstablegetelem{0}{token_reduction}\of{\openingcrosshostdata}\edef\OpenClawOpeningTokRed{\pgfplotsretval}
\pgfplotstablegetelem{1}{token_reduction}\of{\openingcrosshostdata}\edef\OpenCodeOpeningTokRed{\pgfplotsretval}
\pgfplotstablegetelem{0}{legacy_avg_tok_m}\of{\openingbardata}\edef\OpenClawLegacyAvgTok{\pgfplotsretval}
\pgfplotstablegetelem{0}{cs_avg_tok_m}\of{\openingbardata}\edef\OpenClawCsAvgTok{\pgfplotsretval}
\pgfplotstablegetelem{0}{legacy_label}\of{\openingbardata}\edef\OpenClawLegacyBarLabel{\pgfplotsretval}
\pgfplotstablegetelem{0}{cs_label}\of{\openingbardata}\edef\OpenClawCsBarLabel{\pgfplotsretval}
\pgfplotstablegetelem{1}{legacy_avg_tok_m}\of{\openingbardata}\edef\OpenCodeLegacyAvgTok{\pgfplotsretval}
\pgfplotstablegetelem{1}{cs_avg_tok_m}\of{\openingbardata}\edef\OpenCodeCsAvgTok{\pgfplotsretval}
\pgfplotstablegetelem{1}{legacy_label}\of{\openingbardata}\edef\OpenCodeLegacyBarLabel{\pgfplotsretval}
\pgfplotstablegetelem{1}{cs_label}\of{\openingbardata}\edef\OpenCodeCsBarLabel{\pgfplotsretval}

\def\csPromptEnd{2.0}
\def\csRetrievalEnd{4.2}
\def\csOperatingX{4.55}
\def\csSufficientEnd{6.8}
\def\csBaselineY{0.08}
\def\csPlateauY{0.91}

\begin{figure}[tbp]
\centering
{\footnotesize\rmfamily\csMiniLegend}\par\vspace{0.35em}

\begin{minipage}[t]{0.49\textwidth}
\centering
{\footnotesize\rmfamily\textbf{(A) Prevailing code agents}}\par\vspace{0.25em}
\begin{tikzpicture}[x=1cm,y=1cm]
	\begin{pgfonlayer}{background}
		\node [style=wf panel] (panelA) at (0, 0) {};
	\end{pgfonlayer}
	\begin{pgfonlayer}{nodelayer}
		\node [style=wf process, minimum width=2.50cm, minimum height=0.56cm] (taskA) at (3.8, 5.03) {User task\\{\scriptsize issue + prompt}};
		\node [style=wf process, minimum width=6.60cm, minimum height=0.86cm] (acqA) at (3.8, 3.93) {};
		\node [style=wf label, anchor=north west] at ($(acqA.north west)+(0.14,-0.10)$) {Native repository tools};
		\node [style=wf tool raw, minimum width=0.98cm, minimum height=0.30cm] (toolA1) at (2.15, 3.78) {\texttt{Grep}};
		\node [style=wf tool raw, minimum width=0.98cm, minimum height=0.30cm] (toolA2) at (3.80, 3.78) {\texttt{Read}};
		\node [style=wf tool raw, minimum width=0.98cm, minimum height=0.30cm] (toolA3) at (5.45, 3.78) {\texttt{Bash}};
		\node [style=wf process, minimum width=6.60cm, minimum height=0.86cm] (ctxA) at (3.8, 2.68) {};
		\node [style=wf label, anchor=north west] at ($(ctxA.north west)+(0.14,-0.10)$) {Unfiltered context};
		\node [style=wf micro, anchor=north east] at ($(ctxA.north east)+(-0.14,-0.10)$) {low signal-to-noise};
		\draw[draw=csraw!45, line width=0.4pt, rounded corners=1.5pt] (1.10,2.46) rectangle (6.50,2.72);
		\draw[draw=cssniper!70!black, fill=cssniper!35, rounded corners=1pt] (1.12,2.50) rectangle ++(2.60,0.18);
		\node[wf micro, text=cssniper!55!black, inner sep=0pt] at (2.42,2.59) {\texttt{path.py:1--110}};
		\draw[draw=csraw!70!black, fill=csraw!35, rounded corners=1pt] (3.72,2.50) rectangle ++(1.80,0.18);
		\draw[draw=csnoise!70!black, fill=csnoise!35, rounded corners=1pt] (5.52,2.50) rectangle ++(0.96,0.18);
		\node[wf micro, anchor=east, text=csnoise!60!black, inner sep=0pt] at (6.48,2.35) {context window: \textbf{full}};
		\node [style=wf process, minimum width=6.60cm, minimum height=0.50cm] (agentA) at (3.8, 1.61) {Host agent: reason $\rightarrow$ edit $\rightarrow$ test};
		\node [style=wf process warn, minimum width=6.60cm, minimum height=0.58cm] (outA) at (3.8, 0.68) {On failure: reread similar files;\\duplicate context keeps growing};
	\end{pgfonlayer}
	\begin{pgfonlayer}{edgelayer}
		\draw [style=wf arrow] (taskA) to (acqA);
		\draw [style=wf arrow] (acqA) to (ctxA);
		\draw [style=wf arrow] (ctxA) to (agentA);
		\draw [style=wf arrow] (agentA) to (outA);
	\end{pgfonlayer}
\end{tikzpicture}
\end{minipage}\hfill
\begin{minipage}[t]{0.49\textwidth}
\centering
{\footnotesize\rmfamily\textbf{(B) ContextSniper}}\par\vspace{0.25em}
\begin{tikzpicture}[x=1cm,y=1cm]
	\begin{pgfonlayer}{background}
		\node [style=wf panel] (panelB) at (0, 0) {};
	\end{pgfonlayer}
	\begin{pgfonlayer}{nodelayer}
		\node [style=wf process, minimum width=2.50cm, minimum height=0.56cm] (taskB) at (3.8, 5.03) {User task\\{\scriptsize issue + prompt}};
		\node [style=wf process system, minimum width=6.60cm, minimum height=0.86cm] (acqB) at (3.8, 3.93) {};
		\node [style=wf label, text=csblue!55!black, anchor=north west] at ($(acqB.north west)+(0.14,-0.10)$) {ContextSniper insertion};
		\node [style=wf micro, text=csnoise!65!black, anchor=north east] at ($(acqB.north east)+(-0.14,-0.10)$) {suppress noisy outputs};
		\node [style=wf tool system, minimum width=1.55cm, minimum height=0.30cm] (toolB1) at (2.15, 3.78) {\texttt{search\_code}};
		\node [style=wf tool raw, minimum width=0.98cm, minimum height=0.30cm] (toolB2) at (3.80, 3.78) {\texttt{Read}};
		\node [style=wf tool raw, minimum width=0.98cm, minimum height=0.30cm] (toolB3) at (5.45, 3.78) {\texttt{Bash}};
		\node [style=wf process, minimum width=6.60cm, minimum height=0.86cm] (ctxB) at (3.8, 2.68) {};
		\node [style=wf label, anchor=north west] at ($(ctxB.north west)+(0.14,-0.10)$) {Compact evidence packet};
		\node [style=wf micro, anchor=north east] at ($(ctxB.north east)+(-0.14,-0.10)$) {high signal-to-noise};
		\draw[draw=csraw!45, line width=0.4pt, rounded corners=1.5pt] (1.10,2.46) rectangle (6.50,2.72);
		\draw[draw=cssniper!70!black, fill=cssniper!35, rounded corners=1pt] (1.12,2.50) rectangle ++(1.70,0.18);
		\node[wf micro, text=cssniper!55!black, inner sep=0pt] at (1.97,2.59) {\texttt{path.py:42--58}};
		\draw[draw=csraw!70!black, fill=csraw!35, rounded corners=1pt] (2.82,2.50) rectangle ++(0.44,0.18);
		\node[wf micro, anchor=east, text=csblue!60!black, inner sep=0pt] at (6.48,2.35) {context window: \textbf{headroom remaining}};
		\node [style=wf process, minimum width=6.60cm, minimum height=0.50cm] (agentB) at (3.8, 1.61) {Host agent: reason $\rightarrow$ edit $\rightarrow$ test};
		\node [style=wf process system, minimum width=6.60cm, minimum height=0.58cm] (outB) at (3.8, 0.68) {Patch submitted;\\no redundant rereads};
	\end{pgfonlayer}
	\begin{pgfonlayer}{edgelayer}
		\draw [style=wf arrow] (taskB) to (acqB);
		\draw [style=wf system arrow] (acqB) to (ctxB);
		\draw [style=wf arrow] (ctxB) to (agentB);
		\draw [style=wf system arrow] (agentB) to (outB);
	\end{pgfonlayer}
\end{tikzpicture}
\end{minipage}

\vspace{0.8em}

\begin{minipage}[b]{0.62\textwidth}
\centering
{\footnotesize\rmfamily\textbf{(C) Useful-evidence operating point}}\par\vspace{0.25em}
\begin{tikzpicture}
\begin{axis}[
    csfig,
    width=\textwidth, height=2.55in,
    xmin=0, xmax=10, ymin=0, ymax=1.12,
    axis lines=left,
    axis line style={-{Stealth[length=2.2mm]}},
    xlabel={}, ylabel={Usefulness for repair},
    xtick=\empty, ytick=\empty, clip=false,
    legend style={at={(rel axis cs:0.5,1.02)}, anchor=south, legend columns=2,
        /tikz/every even column/.append style={column sep=8pt}},
]
\fill[csprompt, opacity=0.85] (axis cs:0,0) rectangle (axis cs:\csPromptEnd,1.12);
\fill[csband, opacity=0.55] (axis cs:\csPromptEnd,0) rectangle (axis cs:\csRetrievalEnd,1.12);
\fill[cssufficient, opacity=0.55] (axis cs:\csRetrievalEnd,0) rectangle (axis cs:\csSufficientEnd,1.12);
\fill[cswaste, opacity=0.62] (axis cs:\csSufficientEnd,0) rectangle (axis cs:10,1.12);
\draw[csraw!35, dashed, thin] (axis cs:\csPromptEnd,0) -- (axis cs:\csPromptEnd,1.12);
\draw[csraw!35, dashed, thin] (axis cs:\csRetrievalEnd,0) -- (axis cs:\csRetrievalEnd,1.12);
\draw[csraw!35, dashed, thin] (axis cs:\csSufficientEnd,0) -- (axis cs:\csSufficientEnd,1.12);
\node[anchor=north, inner sep=0pt] at (axis cs:1.0,-0.04) {%
  \begin{tabular}{@{}c@{\hspace{0.32em}}c@{}}%
    {\scriptsize\rmfamily\textcolor{csviolet!75!black}{%
      \begin{tabular}{@{}c@{}}task\\instruction\end{tabular}}}%
    & {\begin{tikzpicture}[baseline=-0.55ex,x=1cm,y=1cm]%
        \draw[csviolet,line width=1.3pt,line cap=round] (-0.10,0.10)--(0.05,-0.05);
        \fill[csviolet] (0.05,-0.05)--(0.12,-0.12)--(0.115,-0.02)--cycle;
        \draw[csviolet,line width=0.5pt,line cap=round] (-0.13,0.13)--(-0.07,0.07);
      \end{tikzpicture}}%
  \end{tabular}};
\node[anchor=north, inner sep=0pt] at (axis cs:3.1,-0.04) {%
  \begin{tabular}{@{}c@{\hspace{0.32em}}c@{}}%
    {\scriptsize\rmfamily\textcolor{csblue!65!black}{%
      \begin{tabular}{@{}c@{}}evidence\\search\end{tabular}}}%
    & {\begin{tikzpicture}[baseline=-0.55ex,x=1cm,y=1cm]%
        \draw[csblue,line width=0.6pt] (0,0) circle (0.12);
        \draw[csblue,line width=0.6pt] (0,0) circle (0.065);
        \fill[csblue] (0,0) circle (0.022);
      \end{tikzpicture}}%
  \end{tabular}};
\node[anchor=north, inner sep=0pt] at (axis cs:5.5,-0.04) {%
  \begin{tabular}{@{}c@{\hspace{0.32em}}c@{}}%
    {\scriptsize\rmfamily\textcolor{cssniper!50!black}{%
      \begin{tabular}{@{}c@{}}sufficient\\evidence\end{tabular}}}%
    & {\begin{tikzpicture}[baseline=-0.55ex,x=1cm,y=1cm]%
        \draw[cssniper,fill=cssniper!12,line width=0.5pt,rounded corners=0.04cm] (-0.12,-0.02) rectangle (0.12,0.13);
        \fill[cssniper!12] (-0.05,-0.015)--(-0.02,-0.10)--(0.02,-0.015)--cycle;
        \draw[cssniper,line width=0.5pt,line join=round] (-0.05,-0.02)--(-0.02,-0.10)--(0.02,-0.02);
        \draw[cssniper,line width=0.8pt,line cap=round,line join=round] (-0.06,0.055)--(-0.015,0.015)--(0.07,0.095);
      \end{tikzpicture}}%
  \end{tabular}};
\node[anchor=north, inner sep=0pt] at (axis cs:8.4,-0.04) {%
  \begin{tabular}{@{}c@{\hspace{0.32em}}c@{}}%
    {\scriptsize\rmfamily\textcolor{csnoise!65!black}{%
      \begin{tabular}{@{}c@{}}noisy\\context\end{tabular}}}%
    & {\begin{tikzpicture}[baseline=-0.55ex,x=1cm,y=1cm]%
        \draw[csnoise,fill=csnoise!10,line width=0.5pt] (0,0) circle (0.13);
        \draw[csnoise,line width=0.5pt,line cap=round] (-0.085,0.05)--(-0.04,0.005) (-0.085,0.005)--(-0.04,0.05);
        \draw[csnoise,line width=0.5pt,line cap=round] (0.04,0.05)--(0.085,0.005) (0.04,0.005)--(0.085,0.05);
        \draw[csnoise,line width=0.6pt,line cap=round] (-0.06,-0.06) .. controls (-0.02,-0.03) and (0.02,-0.09) .. (0.06,-0.06);
      \end{tikzpicture}}%
  \end{tabular}};
\draw[csraw, dashed, semithick] (axis cs:\csOperatingX,\csPlateauY) -- (axis cs:10,\csPlateauY);
\addplot[csblue, line width=1.8pt, smooth, forget plot] coordinates {
 (0,\csBaselineY)(\csPromptEnd,\csBaselineY)};
\addplot[csraw, very thick, smooth, forget plot] coordinates {
 (0,\csBaselineY)(\csPromptEnd,\csBaselineY)};
\addplot[csblue, line width=1.8pt, smooth] coordinates {
 (\csPromptEnd,\csBaselineY)
 (2.35,0.12)(2.9,0.29)(3.5,0.60)(3.95,0.84)(\csOperatingX,\csPlateauY)};
\addlegendentry{ContextSniper}
\addplot[csraw, very thick, smooth] coordinates {
 (\csPromptEnd,\csBaselineY)
 (2.55,0.11)(3.3,0.20)(4.1,0.33)(4.9,0.48)(5.8,0.64)(6.5,0.78)(\csSufficientEnd,0.81)
 (7.5,0.80)(8.3,0.69)(9.1,0.54)(10,0.40)};
\addlegendentry{Raw access (\texttt{grep}/\texttt{ls})}
\addplot[only marks, mark=star, mark size=3.8pt, csblue, forget plot]
    coordinates {(\csOperatingX,\csPlateauY)};
\node[font=\scriptsize\rmfamily, anchor=south, text=csblue!55!black]
    at (axis cs:\csOperatingX,0.965) {operating point};
\node[font=\scriptsize\rmfamily, text=csink, anchor=south east]
    at (axis cs:9.95,0.025) {Context cost (tokens)};
\end{axis}
\end{tikzpicture}
\end{minipage}\hfill
\begin{minipage}[b]{0.35\textwidth}
\centering
{\footnotesize\rmfamily\textbf{Avg.\ tokens per task}}\par
{\scriptsize\rmfamily\itshape lower is better}\par\vspace{0.25em}
\begin{tikzpicture}
\begin{axis}[
    csfig,
    width=\textwidth, height=2.35in,
    ybar=1.5pt, bar width=8pt,
    symbolic x coords={OpenClaw,OpenCode}, xtick=data,
    xticklabels={OpenClaw,OpenCode},
    ymin=0, ymax=2.22, ytick={0,0.5,1.0,1.5,2.0}, yticklabels={0,0.5M,1M,1.5M,2M},
    tick label style={font=\footnotesize\rmfamily, /pgf/number format/.cd, fixed},
    enlarge x limits=0.34,
    legend style={at={(0.5,1.06)}, anchor=south, legend columns=2,
        /tikz/every even column/.append style={column sep=6pt}},
    clip=false,
]
\addplot[fill=csraw!55, draw=csraw!80] table[x=host, y=legacy_avg_tok_m, col sep=comma] {resource/opening-figure-bars.csv};
\addplot[fill=csblue, draw=csblue!70!black] table[x=host, y=cs_avg_tok_m, col sep=comma] {resource/opening-figure-bars.csv};
\legend{Baseline, ContextSniper}
\node[font=\scriptsize\rmfamily, anchor=south] at (axis cs:OpenClaw,\OpenClawLegacyAvgTok) {\OpenClawLegacyBarLabel};
\node[font=\scriptsize\rmfamily, anchor=south] at (axis cs:OpenClaw,\OpenClawCsAvgTok) {\OpenClawCsBarLabel};
\node[font=\scriptsize\bfseries\rmfamily, text=csblue!55!black] at (axis cs:OpenClaw,1.66) {$-\OpenClawOpeningTokRed\%$};
\node[font=\scriptsize\rmfamily, anchor=south] at (axis cs:OpenCode,\OpenCodeLegacyAvgTok) {\OpenCodeLegacyBarLabel};
\node[font=\scriptsize\rmfamily, anchor=south] at (axis cs:OpenCode,\OpenCodeCsAvgTok) {\OpenCodeCsBarLabel};
\node[font=\scriptsize\bfseries\rmfamily, text=csblue!55!black] at (axis cs:OpenCode,2.12) {$-\OpenCodeOpeningTokRed\%$};
\end{axis}
\end{tikzpicture}
\end{minipage}
\caption{ContextSniper turns noisy repository context into compact repair evidence. (A) and (B) share the same host-agent workflow; only the context-acquisition module differs. (A) Native \texttt{Grep}, \texttt{Read}, and \texttt{Bash} outputs accumulate into an unfiltered prompt context, and failed attempts reread similar files. (B) A ContextSniper insertion layer replaces broad discovery with \texttt{search\_code} excerpts and gates long \texttt{Read}/\texttt{Bash} output before the same host agent assembles a compact prompt. (C) A conceptual illustration, not measured data, of the intended operating point: context acquisition passes through four stages, instruction, retrieval and edits, sufficient evidence, and context flooding, with ContextSniper reaching the operating point early and stopping while raw \texttt{grep}/\texttt{ls} access keeps spending tokens through redundant reads and only loses accuracy once context floods. The bars give the measured average token use per task (Table~\ref{tab:main-results}).}
\label{fig:contextsniper-overview}
\end{figure}
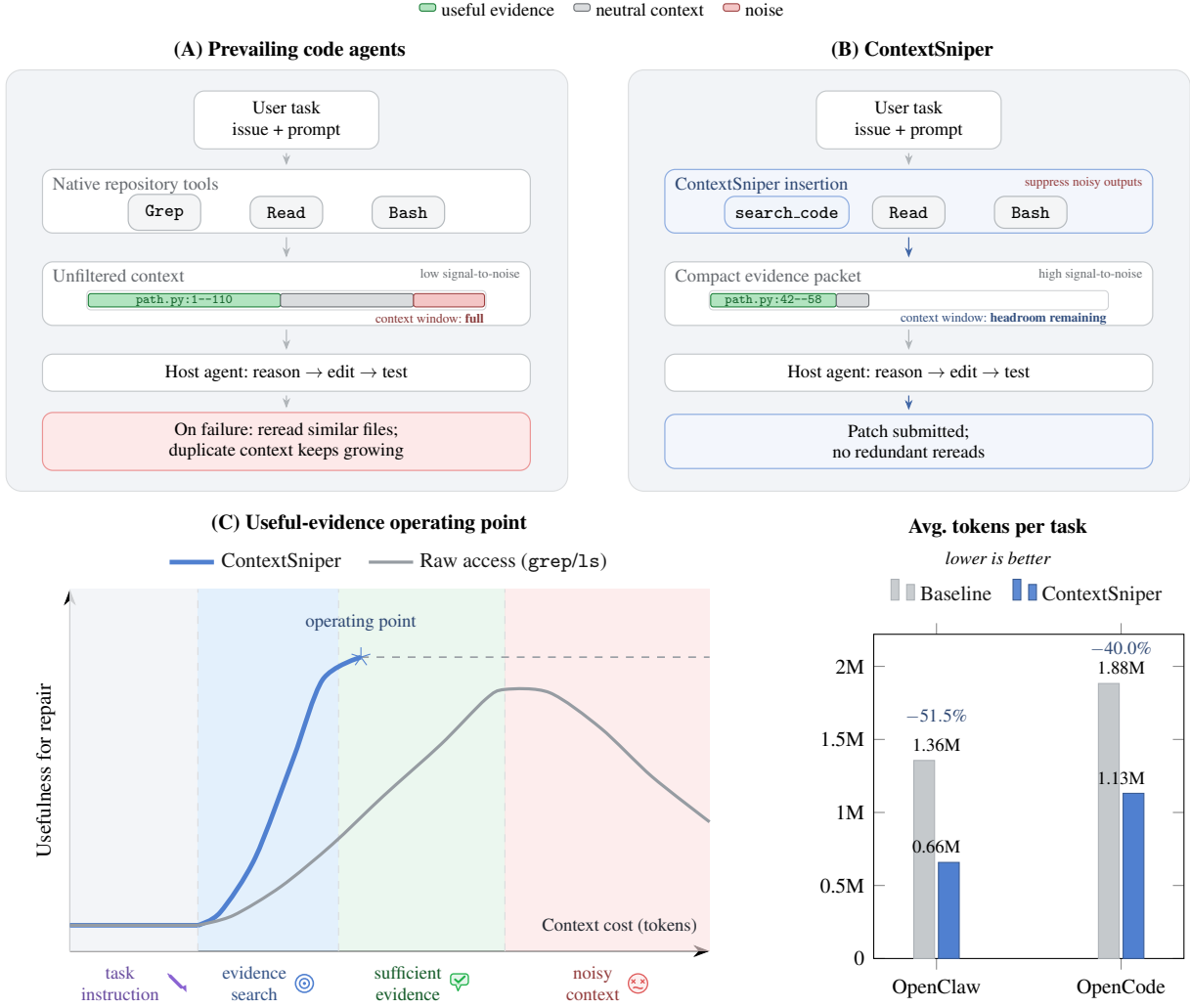

The intended operating point is shown in Figure~\ref{fig:contextsniper-overview}C. More context can help when it adds missing evidence, but after the agent has enough information, additional files and logs mostly add cost and distraction. ContextSniper therefore does not try to maximize context size or compress everything uniformly. It tries to keep the agent near the useful-evidence region: enough code and runtime evidence to localize and repair the issue, but little enough surrounding noise to reduce tokens, turns, and confusion. Rather than replacing the host coding agent, ContextSniper targets the context-access layer between the agent and the repository. It maintains local code memory, retrieves compact task-relevant code, suppresses noisy read and command outputs, and synchronizes memory with repository edits, so the prompt carries the current evidence packet rather than the entire exploration history.

\subsection{Problem Statement}
Repository-level program repair requires an agent to connect an issue description to the relevant implementation, tests, interfaces, and execution behavior of a codebase \parencite{jimenez2024swebench}. This is difficult because the necessary evidence is rarely isolated in one function. It may be distributed across multiple files, imported dependencies, class hierarchies, configuration logic, and failing tests. Before an agent can generate a correct patch, it must first localize the code that matters, a bottleneck also targeted by localization and repository-graph methods \parencite{xia2024agentless,yang2025kgcompass,wang2025repolocmemory}.

Existing agents usually acquire this context through raw repository tools. They read files directly, run search commands, inspect command outputs, and then decide what to read next \parencite{yang2024sweagent,zhang2024autocoderover}. This workflow is general, but it treats context accumulation as the default path to evidence. It provides little memory across repeated exploration steps, little control over the amount of irrelevant output entering the context window, and limited support for keeping retrieved context aligned with files that change during the repair process. Recent codebase-memory and graph-based systems address parts of this problem by giving agents structured repository access, but context selection and runtime output control remain important practical concerns \parencite{ouyang2024repograph,vogel2026codebasememory}.

This paper studies the following problem: how can existing repository-level code agents transform raw repository and runtime context into compact repair evidence while leaving the agent's core reasoning and editing loop intact? The problem has two coupled objectives. The first is to reduce waste: fewer irrelevant tokens, fewer repeated exploration turns, lower cost, and shorter latency. The second is to reduce confusion: cleaner evidence packets that preserve code localization, patch reasoning, and validation quality while removing surrounding noise.

ContextSniper connects to three lines of prior work that target the same context bottleneck from different angles: agent tools and agent-computer interfaces for repository-level software engineering, prompt and retrieval compression, and output-side filtering for agent tools. Agent-tool work shapes how models navigate, search, and edit repositories, but the active context it produces remains tightly coupled to raw exploration output. Prompt- and retrieval-compression methods operate on text or chunks already selected for a prompt, after the selection decision has been made. Output-filtering systems trim noisy tool output but do not maintain a synchronized code memory across a repair run. ContextSniper is distinguished from all three: it gates raw repository and runtime output at the point where it would otherwise enter the model's context, while keeping that gating synchronized with a task-time code memory, rather than compressing an already-assembled prompt or replacing the host agent's tool loop. We position ContextSniper against specific systems in each of these three lines in Section~\ref{sec:background}.

\subsection{Contributions}
We make the following contributions:

\begin{itemize}
  \item We introduce ContextSniper, AntTrail's code-repair module. It sits between a host agent and a target repository, converting noisy repository access into compact evidence packets while leaving the agent's core reasoning and editing loop intact.
  \item We propose a two-family memory hierarchy backed by AGFS (Agent File System), in which code memory and action memory are each indexed as three views: \texttt{L0} compact abstract, \texttt{L1} structured overview, and \texttt{L2} full content.
  \item We develop adaptive top-\(k\) retrieval that fuses semantic embeddings, BM25 lexical scoring, ctags-style symbolic metadata (symbol names, kinds, and locations), and graph relations through weighted reciprocal rank fusion (RRF), with a ripgrep-based text-search fallback \parencite{burntsushi2026ripgrep} when the index misses a query.
  \item We add an intention-aware context gate and a two-phase memory-repository synchronization that snipe noisy tool output and keep local code memory aligned with the working tree after agent edits.
  \item We evaluate ContextSniper on SWE-bench Lite with OpenClaw as the host agent, comparing its baseline against ContextSniper in terms of token usage, tool behavior, and repair outcomes \parencite{jimenez2024swebench,openclaw2026github}. ContextSniper reduces total token use by 51.5\%, with a comparable submitted-resolution rate.
\end{itemize}

\section{Background and Related Work}
\label{sec:background}
\fontsize{10}{12}\selectfont

\subsection{Repository-Level Program Repair}
Repository-level program repair asks a model or agent to modify an existing codebase in response to a natural-language issue report, rather than solving an isolated programming exercise. SWE-bench made this setting a standard evaluation target by collecting real GitHub issues and pull requests from Python repositories; its tasks require models to inspect a repository, understand issue-specific behavior, and coordinate edits across files, tests, and execution environments \parencite{jimenez2024swebench}. This benchmark shifted automated repair from single-function generation toward interactive repository work, where finding the right evidence can be as important as generating the final patch.

Recent systems can be separated along four dimensions: how autonomous the model is, how localization is performed, whether memory persists across interactions or tasks, and whether raw repository or runtime output is filtered before entering the prompt. SWE-agent studies the agent-computer interface, showing that repository navigation, file editing, and test execution tools can strongly affect software-engineering agent behavior \parencite{yang2024sweagent}. AutoCodeRover uses a more software-engineering-oriented autonomous workflow, combining LLM reasoning with structured code search and test-aware fault localization \parencite{zhang2024autocoderover}. Agentless, by contrast, separates localization, repair, and patch validation into a simpler pipeline rather than giving the model a fully autonomous tool loop \parencite{xia2024agentless}. Table~\ref{tab:repair-positioning} summarizes the resulting boundary.

\begin{table}[t]
\centering
\footnotesize
\setlength{\tabcolsep}{3pt}
\caption{Positioning ContextSniper relative to repository-level repair systems.}
\label{tab:repair-positioning}
\begin{tabular}{L{0.19\linewidth}L{0.22\linewidth}L{0.25\linewidth}L{0.26\linewidth}}
\toprule
Family & Examples & Primary optimization target & Context-access boundary \\
\midrule
Benchmark setting & SWE-bench \parencite{jimenez2024swebench} & Real issue-to-patch tasks over full repositories. & Defines the task and validation setting, but does not prescribe how evidence should be retrieved or bounded. \\
\addlinespace
Autonomous repair agents & SWE-agent and AutoCodeRover \parencite{yang2024sweagent,zhang2024autocoderover} & Tool use, repository navigation, search, edit, test, and agent-computer interface design. & Context selection is largely coupled to the host agent's exploratory tool loop and the raw outputs that loop observes. \\
\addlinespace
Pipeline-style repair & Agentless \parencite{xia2024agentless} & A simpler decomposition into localization, repair, and validation. & Localization is a pipeline stage; reusable task-time memory, runtime-output gating, and edit synchronization are not the main abstraction. \\
\addlinespace
Context-access layer & ContextSniper (AntTrail) & Evidence retrieval, budgeting, filtering, and synchronization for an existing host agent. & The repair agent remains responsible for reasoning and editing; the context layer controls which repository and runtime evidence enters the prompt. \\
\bottomrule
\end{tabular}
\end{table}

This distinction is central to the paper's contribution. ContextSniper is not a new standalone repair agent and does not replace the host model's reasoning loop. Instead, it targets the context-access layer used by repair agents: how repository evidence is retrieved, bounded, selected, filtered, and kept synchronized with edits. This makes it complementary to autonomous systems such as SWE-agent and AutoCodeRover and pipeline-style systems such as Agentless.

\subsection{Code Memory and Agentic Memory}
Retrieval-augmented generation introduced a general pattern for combining a parametric language model with explicit non-parametric memory, allowing retrieved passages to condition generation without retraining the model \parencite{lewis2020rag}. Software-engineering agents extend this idea because useful knowledge may include source code, documentation, prior fixes, execution traces, project conventions, and failure histories. The important distinction for this paper is not only whether memory exists, but what scope the memory covers and when it is updated.

One line of work stores cross-task repair experience. SWE-Exp distills concise knowledge from prior agent trajectories into an experience bank for issue resolution \parencite{chen2025sweexp}. ExpeRepair uses episodic and semantic memories to retrieve concrete repair demonstrations and higher-level repair insights \parencite{mu2025experepair}. MemGovern expands the source of experience to governed human issue-tracking data, aiming to move agents beyond local context alone \parencite{wang2026memgovern}. A second line studies memory granularity: structurally aligned subtask-level memory argues that whole-instance memories can be too coarse and stores or retrieves experience at the level of functional subtasks \parencite{shen2026subtaskmemory}. A third line studies domain adaptation, as in MEMCoder, which evolves memory for private-library-oriented code generation where the model must learn internal API usage patterns and constraints over time \parencite{li2026memcoder}.

These systems show that memory is becoming central to software agents, but their main emphasis is experience reuse across tasks, histories, or domains. ContextSniper focuses on task-time local code memory. Its memory is built from the current repository snapshot, queried during the current repair run, refreshed as files change, and used to control not only code retrieval but also what read, search, and shell outputs enter the active context. This narrower scope is intentional: it addresses repeated exploration and noisy context within a single issue-resolution process, while leaving room for future integration with cross-task experience memory.

\subsection{Repository Structure, Localization, and Context Management}
Repository-level repair depends on localization before generation. If the agent reads the wrong files or misses the right symbol, later patch reasoning is likely to fail regardless of model size. Prior work therefore uses repository structure, history, or graphs to narrow the search space. RepoGraph provides a repository-level code graph as a plug-in module for AI software-engineering systems \parencite{ouyang2024repograph}. KGCompass uses repository-aware knowledge graphs over artifacts such as issues, pull requests, files, classes, and functions to guide fault localization and repair \parencite{yang2025kgcompass}. Repository-memory localization work builds non-parametric memory from historical commits and linked issues, showing that a project's change history can help agents identify likely fix locations \parencite{wang2025repolocmemory}. Learning to Commit similarly studies online repository memory for generating pull requests that respect project-specific conventions and architectural patterns \parencite{li2026learningcommit}.

Persistent codebase memory systems address a closely related practical bottleneck: LLM coding agents often explore repositories through repeated file reads and broad search commands. Codebase-Memory constructs a Tree-Sitter-based knowledge graph exposed through the Model Context Protocol and reports large token reductions for codebase exploration tasks \parencite{vogel2026codebasememory}. This line of work supports the central premise of ContextSniper: repository agents need structured access to code, not only larger context windows. ContextSniper differs by treating the final prompt as a budgeted evidence packet that must preserve source provenance, filter runtime noise, and remain synchronized with the repository after edits.

Long-context models do not remove this selection problem. Empirical work on long-context language models shows that performance can degrade when relevant information is buried in the middle of long inputs, even for models designed to accept long contexts \parencite{liu2024lostmiddle}. This result is not a direct evaluation of repository repair, but it motivates caution: adding more files, logs, and command output is not equivalent to making repair evidence easier to use. The gap left by prior work is therefore a context-management problem with four requirements: select task-relevant local repository evidence under a budget, preserve provenance so the host agent can inspect and edit the source, suppress low-value runtime output without losing diagnostic anchors, and keep retrieved memory aligned with changes made during the repair run.

\subsection{Prompt and Retrieval Compression}
A separate line of work optimizes prompt and retrieval compression once content has already been selected for a prompt. \textsc{LLMLingua} and \textsc{LongLLMLingua} drop low-information tokens via a small LM, achieving up to 20$\times$ compression with small quality loss and explicit mitigation of the lost-in-the-middle failure mode \parencite{jiang2023llmlingua,jiang2024longllmlingua}. \textsc{RECOMP} inserts a learned compressor between retrieval and the LM and ships both extractive and abstractive variants for retrieval-augmented generation (RAG) settings \parencite{xu2023recomp}. \textsc{COMPACT} makes the compression rate query-dependent, deciding per retrieved document how aggressively to shorten it \parencite{yoon2024compact}. \textsc{Selective Context} filters tokens by self-information \parencite{li2023selectivecontext}, while \textsc{AutoCompressor} trains the model to emit summary vectors that compress earlier context into the KV cache \parencite{chevalier2023autocompressor}. Open-source engineering tools in this layer include \textsc{ctxbudgeter}, which compiles, audits, governs, and visualizes context packs before model calls with token budgets, policy checks, provenance, and cache planning \parencite{ctxbudgeter2026github}. These methods operate on text or retrieved chunks already selected for a prompt; ContextSniper operates one layer earlier for repository repair, deciding which tool outputs and code excerpts are allowed to enter that text at all.

\subsection{Agent-Tool Output Filtering}
A third line filters agent-tool outputs before they enter the model context. \textsc{RTK} rewrites shell commands through a Rust CLI proxy and reports 60--90\% token savings on common development commands by filtering, grouping, truncating, and deduplicating outputs \parencite{rtk2026github}. \textsc{Headroom} generalizes this idea beyond shell commands to tool outputs, logs, files, RAG chunks, and conversation history through a library, proxy, and MCP server, reporting 60--95\% fewer tokens on real agent workloads while retaining local-first operation and reversible retrieval of originals \parencite{headroom2026github}. \textsc{Bearing} addresses context growth at the workflow level: it separates planning from execution, runs tasks in fresh agent processes, and passes forward task summaries and relevant-file hints rather than an entire accumulated conversation \parencite{bearing2026github}. ContextSniper is closest to this family, but specializes the output gate for repository-level repair: it combines tool-output filtering with synchronized local code memory, so the host agent receives compact, task-relevant repair evidence without changing its outer workflow. Existing agent tools, prompt compressors, and output filters each improve one link of the evidence chain; to our knowledge, none of them gates raw repository and runtime output for repair while maintaining synchronized code memory. ContextSniper combines these ideas into a single context-access layer and is evaluated against the prevailing raw-access workflow on a repository-level repair benchmark \parencite{jimenez2024swebench}, rather than against an abstract compression ratio on a synthetic long-context task.

\section{ContextSniper: AntTrail's Code-Oriented Memory Module}
\fontsize{10}{12}\selectfont

\begin{figure}[tbp]
\centering
\makebox[\textwidth][c]{\includegraphics[width=1.05\textwidth]{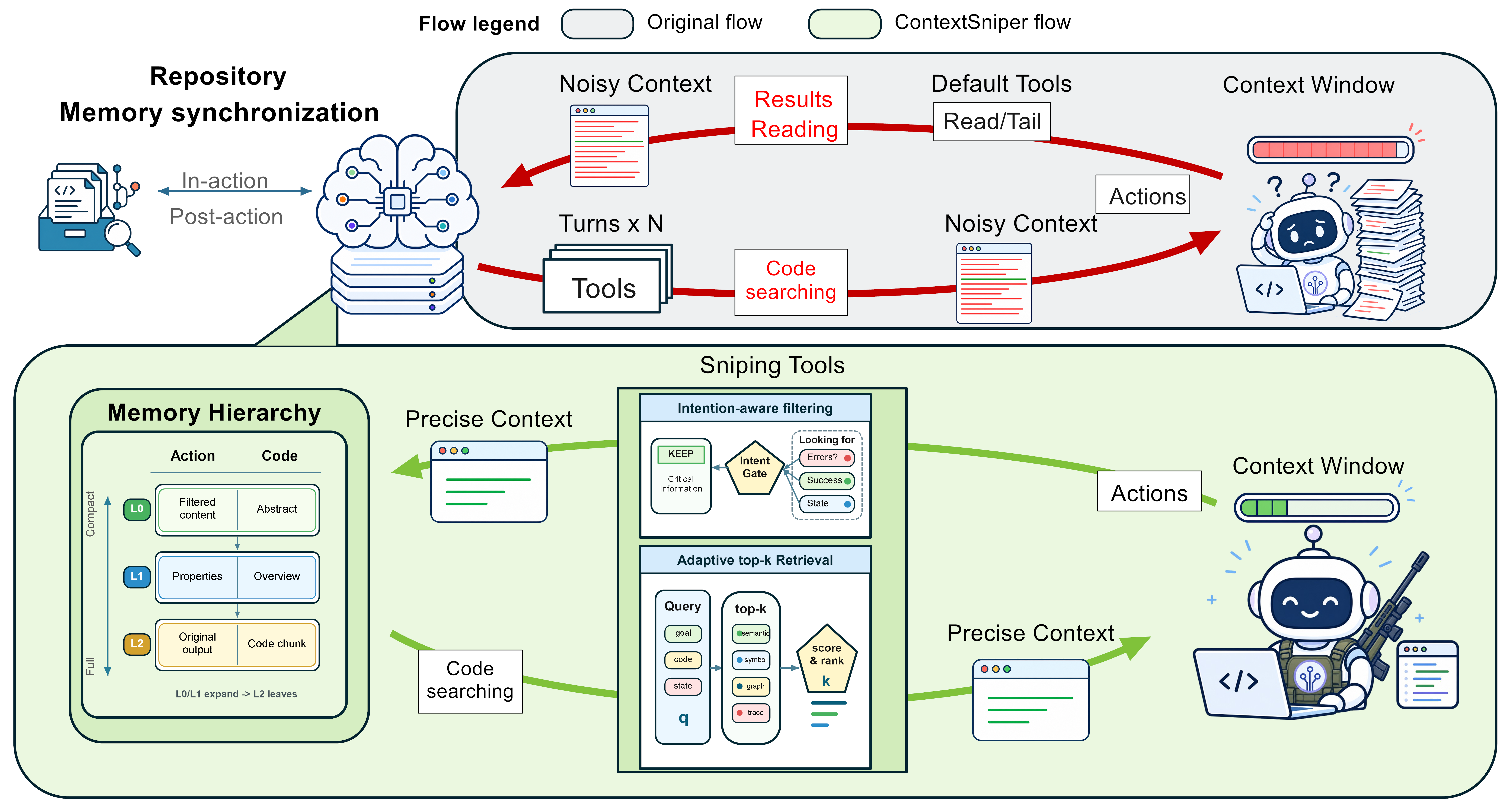}}
\caption{ContextSniper architecture. The upper path shows the baseline interaction loop, where broad search, long file reads, repeated actions, and raw command output accumulate noisy context before reaching the model's context window. The lower path shows the ContextSniper insertion layer. Repository and action evidence are stored in an AGFS-backed memory hierarchy, synchronized with the current working tree, retrieved through hybrid code search, and sniped by intention-aware context gates before compact evidence returns to the same host-agent loop.}
\label{fig:contextsniper-architecture}
\end{figure}

\subsection{System Overview}
Within AntTrail, ContextSniper is the module specialized for repository-level program repair. It is inserted between a host coding agent and a target repository, but it does not replace the agent's planning loop, editor, shell, or validation procedure. Instead, it changes the evidence path. In the baseline path in Figure~\ref{fig:contextsniper-architecture}, the agent explores the repository through broad search, whole-file reads, and command outputs. These outputs often contain the needed repair evidence, but they also carry unrelated functions, repeated logs, and stale exploratory context into the model window. ContextSniper keeps the same outer workflow while changing what the model receives from each interaction.

The core principle is to retrieve broadly but expose narrowly. ContextSniper stores full repository and action evidence outside the prompt, searches that evidence with multiple retrieval signals, and returns only the snippets and runtime facts that are likely to support localization, patch construction, or validation. This is different from generic compression: the system is not trying to summarize everything the agent saw. It is trying to remove unwanted context while preserving source-grounded evidence. Every returned code excerpt keeps its repository path and line region, and every retained runtime signal is tied to the tool call, command, or edit that produced it.

After this overview, the architecture has five implementation components. Section~3.2 describes the AGFS-backed memory hierarchy used to store code and action evidence. Section~3.3 describes synchronization between memory and the working repository. Section~3.4 describes code search and adaptive top-\(k\) retrieval, including semantic embeddings, BM25 lexical scoring, symbol ranking, graph evidence, reciprocal-rank fusion, and ripgrep fallback \parencite{robertson2009bm25,cormack2009rrf,burntsushi2026ripgrep}. Section~3.5 describes the context gate that snipes, or removes low-value regions from, long reads and command outputs before they enter the model prompt. Section~3.6 defines the host-agent boundary; the concrete pilot adapters are described in Section~4.1.

\subsection{Memory Hierarchy}
For code repair, the memory layer is organized into two families: code memory for repository evidence and action memory for evidence produced during the repair run. Both are stored outside the prompt in an AGFS-backed local memory service \parencite{agfs2026github}. AGFS (Agent File System) stores each memory node as a directory with three text views: an abstract, an overview, and full content. ContextSniper indexes these views as three retrieval levels: L0, L1, and L2.

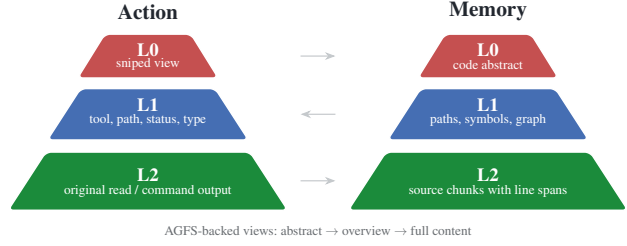
\begin{figure}[tbp]
\centering
\begin{tikzpicture}[x=1cm,y=1cm]
  \tikzset{
    hierarchy/layer/.style={line width=0.55pt, draw=white, rounded corners=2pt},
    hierarchy/title/.style={font=\footnotesize\bfseries\rmfamily, text=csink},
    hierarchy/level/.style={font=\scriptsize\bfseries\rmfamily, text=white, align=center},
    hierarchy/detail/.style={font=\tiny\rmfamily, text=white, align=center},
    hierarchy/caption/.style={font=\tiny\rmfamily, text=csraw!70!black, align=center}
  }

  \node[hierarchy/title] at (-2.25,2.98) {Action};
  \node[hierarchy/title] at (2.25,2.98) {Memory};

  \path[hierarchy/layer, fill=csmark!88!black] (-3.18,2.10) -- (-1.32,2.10) -- (-1.68,2.68) -- (-2.82,2.68) -- cycle;
  \path[hierarchy/layer, fill=csblue!86!black] (-3.58,1.28) -- (-0.92,1.28) -- (-1.36,1.96) -- (-3.14,1.96) -- cycle;
  \path[hierarchy/layer, fill=cssniper!78!black] (-4.10,0.35) -- (-0.40,0.35) -- (-0.90,1.12) -- (-3.60,1.12) -- cycle;

  \node[hierarchy/level] at (-2.25,2.48) {L0};
  \node[hierarchy/detail] at (-2.25,2.28) {sniped view};
  \node[hierarchy/level] at (-2.25,1.75) {L1};
  \node[hierarchy/detail] at (-2.25,1.54) {tool, path, status, type};
  \node[hierarchy/level] at (-2.25,0.82) {L2};
  \node[hierarchy/detail] at (-2.25,0.60) {original read / command output};

  \path[hierarchy/layer, fill=csmark!88!black] (1.32,2.10) -- (3.18,2.10) -- (2.82,2.68) -- (1.68,2.68) -- cycle;
  \path[hierarchy/layer, fill=csblue!86!black] (0.92,1.28) -- (3.58,1.28) -- (3.14,1.96) -- (1.36,1.96) -- cycle;
  \path[hierarchy/layer, fill=cssniper!78!black] (0.40,0.35) -- (4.10,0.35) -- (3.60,1.12) -- (0.90,1.12) -- cycle;

  \node[hierarchy/level] at (2.25,2.48) {L0};
  \node[hierarchy/detail] at (2.25,2.28) {code abstract};
  \node[hierarchy/level] at (2.25,1.75) {L1};
  \node[hierarchy/detail] at (2.25,1.54) {paths, symbols, graph};
  \node[hierarchy/level] at (2.25,0.82) {L2};
  \node[hierarchy/detail] at (2.25,0.60) {source chunks with line spans};

  \draw[cs/arrow, draw=csraw!55] (-0.28,2.39) -- (0.28,2.39);
  \draw[cs/arrow, draw=csraw!55] (0.28,1.62) -- (-0.28,1.62);
  \draw[cs/arrow, draw=csraw!55] (-0.28,0.74) -- (0.28,0.74);
  \node[hierarchy/caption] at (0,0.08) {AGFS-backed views: abstract $\rightarrow$ overview $\rightarrow$ full content};
\end{tikzpicture}
\caption{Two-family L0-L2 memory hierarchy. Action memory stores compact action views, action metadata, and recoverable original tool output. Code memory stores code abstracts, structured routing metadata, and source-grounded chunks.}
\label{fig:memory-hierarchy-pyramid}
\end{figure}

L0 is the compact abstract level: a short model-facing view used for broad recall and, for long reads or shell output, as the returned sniped view. L1 is the structured overview level, storing routing and ranking metadata such as paths, line ranges, symbols, signatures, ctags-style symbol kinds, BM25 lexical documents, graph relations, tool names, command status, and output type \parencite{universalctags2026github}. L2 is the recoverable full-content level. For code memory, L2 contains source-grounded chunks with path and line headers; Python files use an abstract syntax tree (AST) chunker for top-level classes, functions, and async functions, while other files fall back to overlapping line windows. For action memory, L2 stores the original unfiltered read or command output. Thus L0 gives compact recall, L1 gives structured routing, and L2 preserves editable or recoverable source evidence.

\subsection{Memory-Repository Synchronization}
Memory-repository synchronization keeps code memory aligned with the current working tree. A repair run starts from a checked-out repository, but the repository does not stay fixed: the agent edits files, executes tests, reads new regions, and may trigger generated artifacts or logs. If retrieval returns an old version of a function after an edit, the compact context becomes actively harmful. ContextSniper therefore treats memory freshness as part of source grounding.

Synchronization happens in two phases. The first phase runs after agent actions through runtime hooks. When the host agent searches code, reads a file, edits source, or runs a command, ContextSniper records touched paths, returned regions, command status, failure output, and modified files. For edits made through the ContextSniper tool surface, the edit operation validates the target path, applies the source change, and dispatches a memory refresh for the edited file. For long reads and command outputs, the hook writes both the sniped view and the recoverable original into action memory.

The second phase runs at conversation or task boundaries. When a new user prompt arrives, ContextSniper reconciles the indexed state with the working tree, for example by inspecting repository diffs since the previous turn. Changed files are re-chunked or refreshed, deleted snippets are invalidated, and newly relevant regions become available for subsequent searches. This phase catches changes that did not pass through a tool hook and prevents the next repair request from starting from a stale memory image.

The result is a memory hierarchy that is both compact and auditable. Returned snippets cite concrete paths and line regions in the current workspace. Runtime signals cite the command, read, or edit that produced them. Synchronization is therefore not an auxiliary bookkeeping feature; it is the mechanism that makes context sniping trustworthy after the agent starts changing the repository.

\subsection{Code Searching with Adaptive \texorpdfstring{Top-\(k\)}{Top-k} Retrieval}
Code search is the main entry point from the host agent into code memory. Given a focused natural-language or keyword query, \texttt{search\_code} returns a small list of L2 snippets with path and line headers rather than raw grep output. Formally, it computes an evidence packet \(E=\{e_1,\ldots,e_m\}\) from a query \(q\), scope \(S\), memory hierarchy \(M=\{L0,L1,L2\}\), repository state \(R\), requested top-\(k\) value \(k_0\), and context budget \(B\). The adaptive part is the choice of effective return size \(m \leq k^\ast\) and route budgets: traceback-like queries emphasize paths, symbols, and grep terms; behavioral queries allocate more budget to semantic and graph routes; and stateful debugging turns can promote recently touched files from action memory.

The retrieval path is layer aware. A planner assigns the query type and route budgets; seed retrieval searches L0/L1/L2 records; promising L0/L1 nodes expand into L2 leaves; and the final ranker deduplicates by URI, sorts by score, and returns source-grounded L2 evidence. Figure~\ref{fig:adaptive-topk-search} gives the pseudocode.

The hybrid ranker combines semantic embeddings, BM25 lexical documents \parencite{robertson2009bm25}, ctags-style symbol metadata \parencite{universalctags2026github}, and graph relations over imports, calls, containment, and neighboring-file evidence. ContextSniper fuses route rankings with weighted reciprocal rank fusion by default, or weighted score summation when configured \parencite{cormack2009rrf}. If indexed retrieval is weak, it falls back to ripgrep-based workspace search \parencite{burntsushi2026ripgrep}. The important contract is that L0 and L1 help find evidence, but the agent sees precise code regions rather than every file or line that matched the query.

\begin{figure}[tbp]
\centering
\begin{minipage}{0.94\textwidth}
\footnotesize
\hrule height 0.9pt
\vspace{0.4em}
\textbf{Input:} query \(q\), scope \(S\), memory \(M\), state \(R\), action memory \(A\), ceiling \(k_0\), budget \(B\).\\
\textbf{Output:} ranked L2 evidence packet \(E\).

\vspace{0.35em}
\hrule height 0.35pt
\vspace{0.35em}
\begin{tabbing}
\quad\=\quad\=\quad\=\quad\=\kill
\textbf{function} \textsc{SearchCode}(\(q,S,M,R,A,k_0,B\))\\
\> \(intent \leftarrow \textsc{InferIntent}(q,A)\)\\
\> \(k^\ast \leftarrow \min(k_0,\textsc{BudgetToK}(B))\)\\
\> \(routes \leftarrow \textsc{PlanRoutes}(intent,A,k^\ast)\)\\
\> \textbf{for} \(r \in routes\) \textbf{do}\\
\>\> \(C_r \leftarrow \textsc{ExpandToL2}(\textsc{Retrieve}(r,q,S,M,r.budget),M)\)\\
\> \textbf{end for}\\
\> \(F \leftarrow \textsc{Deduplicate}(\textsc{WeightedRRF}(\{C_r\}_{r\in routes},intent),\textit{path},\textit{lineSpan})\)\\
\> \textbf{if} \(\textsc{WeakRecall}(F,S,k^\ast)\) \textbf{then}\\
\>\> \(F \leftarrow \textsc{MergeAndRank}(F,\textsc{Backfill}(\{C_r\},k^\ast),\textsc{Ripgrep}(q,S,R,k^\ast))\)\\
\> \textbf{end if}\\
\> \(E \leftarrow \textsc{Prune}(F,B,k^\ast,\textit{scoreGap},\textit{duplicateText})\)\\
\> \textbf{return} \(\textsc{FillL2Content}(E,M)\)\\
\textbf{end function}
\end{tabbing}
\vspace{0.15em}
\hrule height 0.9pt
\end{minipage}
\caption{Pseudocode for code search with adaptive top-\(k\) retrieval. The query planner controls retrieval depth and route budgets, while the final response remains a compact set of source-grounded L2 snippets.}
\label{fig:adaptive-topk-search}
\end{figure}

\subsection{Intention-Aware Filtering and Context Gate}
The context gate controls what reaches the model after retrieval or tool execution. For file reads, it preserves relevant local definitions, imports, class or function bodies, and nearby code. For test and shell output, it preserves command lines, exit status, assertion messages, tracebacks, failing paths, and concise success signals. For state, it preserves changed files, touched symbols, repository status, and paths that connect the current turn to earlier evidence. Long boilerplate, repeated logs, unrelated file regions, and uninformative tails are removed from the prompt-facing view.

The gate is conservative because removing evidence can be worse than spending extra tokens. When a read or command is shortened, ContextSniper stores the original output in action memory and returns a marker indicating that sniping occurred, so the host agent can request broader context if the compact view is insufficient. Figure~\ref{fig:context-gate} gives the pseudocode.

\begin{figure}[tbp]
\centering
\begin{minipage}{0.94\textwidth}
\footnotesize
\hrule height 0.9pt
\vspace{0.4em}
\textbf{Input:} action \(a\), raw output \(o\), intent query \(q\), state \(R\), action memory \(A\), budget \(B\).\\
\textbf{Output:} prompt-facing view \(v\), sniping marker \(m\), recoverable action-memory record \(u\).

\vspace{0.35em}
\hrule height 0.35pt
\vspace{0.35em}
\begin{tabbing}
\quad\=\quad\=\quad\=\quad\=\kill
\textbf{function} \textsc{ContextGate}(\(a,o,q,R,A,B\))\\
\> \(intent \leftarrow \textsc{InferPreservationIntent}(a,q,A)\)\\
\> \(meta \leftarrow \textsc{ClassifyOutput}(a,o)\)\\
\> \textbf{switch} \(meta.type\) \textbf{do}\\
\>\> \textbf{case} \(\textsc{FileRead}\): \(K \leftarrow \textsc{RequestedRegion}(a,o) \cup \textsc{LocalDefs}(o,intent) \cup \textsc{Neighbors}(o,intent)\)\\
\>\> \textbf{case} \(\textsc{CommandOutput}\): \(K \leftarrow \textsc{CommandStatus}(a,o) \cup \textsc{Failures}(o) \cup \textsc{Validation}(o)\)\\
\>\> \textbf{otherwise}: \(K \leftarrow \textsc{StateSignals}(a,o,R,A)\)\\
\> \textbf{end switch}\\
\> \(v \leftarrow \textsc{RenderCompactView}(K,B)\)\\
\> \(risk \leftarrow \textsc{LowConfidence}(v,intent) \lor \textsc{MissingCriticalSignal}(v,meta)\)\\
\> \(u \leftarrow \textsc{StoreActionMemory}(L0=v,L1=meta,L2=o)\)\\
\> \textbf{if} \(\textsc{Shortened}(v,o) \lor risk\) \textbf{then}\\
\>\> \(m \leftarrow \textsc{SnipedMarker}(u,risk)\)\\
\>\> \(v \leftarrow \textsc{AttachRecoveryHint}(v,m)\)\\
\> \textbf{else}\\
\>\> \(m \leftarrow \textsc{NoSniping}\)\\
\> \textbf{end if}\\
\> \textbf{return} \((v,m,u)\)\\
\textbf{end function}
\end{tabbing}
\vspace{0.15em}
\hrule height 0.9pt
\end{minipage}
\caption{Pseudocode for intention-aware filtering and the context gate. The gate keeps action-relevant evidence, records the recoverable original output, and marks shortened views so the host agent can request broader context when needed.}
\label{fig:context-gate}
\end{figure}

\subsection{Host-Agent Boundary}
ContextSniper is designed as an insertion layer for existing coding agents. The host agent chooses the plan, decides what to edit, and runs validation; the insertion layer controls how repository and runtime evidence are searched, sniped, synchronized, and returned. This boundary allows the mechanism to support a host agent without changing its core reasoning loop. The OpenClaw pilot adapter used in this paper is described in the experimental setup.

\section{Evaluation}
\fontsize{10}{12}\selectfont
\pgfplotstableread[col sep=comma]{resource/summary-scalars.csv}\summarydata
\newcommand{\resultcell}[1]{\pgfplotstablegetelem{0}{#1}\of{\summarydata}\pgfplotsretval}
\pgfplotstableread[col sep=comma]{resource/cross-host-efficiency.csv}\crosshostdata
\pgfplotstablegetelem{0}{token_reduction}\of{\crosshostdata}\edef\OpenClawTokenReduction{\pgfplotsretval}
\pgfplotstablegetelem{0}{cost_reduction}\of{\crosshostdata}\edef\OpenClawCostReduction{\pgfplotsretval}
\pgfplotstablegetelem{1}{token_reduction}\of{\crosshostdata}\edef\OpenCodeTokenReduction{\pgfplotsretval}
\pgfplotstablegetelem{1}{cost_reduction}\of{\crosshostdata}\edef\OpenCodeCostReduction{\pgfplotsretval}
\pgfplotstableread[col sep=comma]{resource/opening-figure-bars.csv}\avgtokdata
\pgfplotstablegetelem{0}{legacy_label}\of{\avgtokdata}\edef\OpenClawLegacyAvgTok{\pgfplotsretval}
\pgfplotstablegetelem{0}{cs_label}\of{\avgtokdata}\edef\OpenClawCsAvgTok{\pgfplotsretval}

\subsection{Experimental Setup}
\begin{table}[H]
\centering
\small
\setlength{\tabcolsep}{4pt}
\caption{Experimental settings. The embedding model is used only in the ContextSniper condition.}
\label{tab:experimental-settings}
\begin{tabular}{llll}
\toprule
Benchmark & Agent & Base model & Embedding model \\
\midrule
SWE-bench Lite & OpenClaw & DeepSeek V4 Flash & \texttt{text-embedding-3-large} \\
SWE-bench Pro & OpenCode (\texttt{build}) & DeepSeek V4 Flash & \texttt{text-embedding-3-small} \\
\bottomrule
\end{tabular}
\end{table}

We evaluate ContextSniper on repository-level program repair in two host-agent settings drawn from the SWE-bench benchmark family \parencite{jimenez2024swebench}. Each study uses 50 experimental instances randomly sampled from its benchmark: SWE-bench Lite for OpenClaw and SWE-bench Pro for OpenCode. For each benchmark instance, the runner checks out the target repository at the benchmark-provided base commit, constructs a task prompt from the issue description, and asks the host agent to generate a source-code patch. Generated patches are then scored by the corresponding benchmark validation harness.

The main comparison is each host agent's baseline versus the ContextSniper pilot implementation. The baseline condition uses the host agent's default repository tools without the ContextSniper insertion layer. ContextSniper adds local code-memory search, compact evidence assembly, context control for long file-read and command-output results, and source editing through the same memory-aware interface.

We evaluate ContextSniper with OpenClaw, an open-source agent setting where ContextSniper exposes its functions through plugin tools and runtime hooks \parencite{openclaw2026github}. The key tool is \texttt{search\_code}, which is presented as the preferred replacement for broad grep and exploratory reads; a paired \texttt{edit\_file} tool validates workspace paths, applies exact source replacements, and refreshes memory for the edited file. Token usage is the primary efficiency metric, including total tokens and, when available, cost and tool/action counts. Validation outcomes are reported cautiously because solve-rate claims require comparable official validation coverage across conditions. The public repository contains the pilot testing scripts and evaluation harnesses used for these experiments, rather than the full AntTrail memory engine.

The OpenCode comparison uses plain OpenCode as the baseline and OpenCode with the full ContextSniper tool bundle as the treatment. Both arms use the \texttt{deepseek/deepseek-v4-flash} model and the autonomous \texttt{build} agent, with web tools disabled. The retained re-runs use OpenCode 1.18.12 and a 900-second timeout per instance. Patches are evaluated with the official SWE-bench Pro Docker harness on \texttt{linux/amd64}, using the local benchmark image set with image pulling disabled and no host-local scoring fallback. Correctness is the proportion of resolved instances over the 50 experimental instances. Resource metrics are also reported for this 50-instance experiment: token consumption is computed from the \texttt{total} field in OpenCode \texttt{step\_finish} events, and rounds are the number of those events.

The reported OpenCode result is a derived comparison rather than a full re-execution of the treatment arm. It retains the 50 randomly sampled experimental instances and incorporates two accepted ContextSniper re-runs. One replacement both changed the verdict to resolved and reduced tokens and rounds; the other was accepted for its resolved verdict despite using more resources than its historical ContextSniper record. We therefore report the adoption provenance when interpreting the aggregate results.

\subsection{Main Results}
Across both host-agent settings, the results support an efficiency claim rather than a correctness-improvement claim. On the \resultcell{openclaw_tasks}-task OpenClaw SWE-bench Lite comparison, both the OpenClaw baseline and ContextSniper completed \resultcell{openclaw_tasks} task runs and submitted \resultcell{openclaw_tasks} patches. ContextSniper reduces average token use from \OpenClawLegacyAvgTok{} to \OpenClawCsAvgTok{} tokens per task, a \resultcell{openclaw_token_reduction}\% reduction in total tokens. It also reduces logged cost by \resultcell{openclaw_cost_reduction}\% and tool/action count by \resultcell{openclaw_action_reduction}\% compared with the OpenClaw baseline. Official validation gives comparable submitted-resolution rates, \resultcell{cs_resolution_rate}\% for ContextSniper versus \resultcell{legacy_resolution_rate}\% for the OpenClaw baseline, but the validation-error imbalance prevents a strong repair-success conclusion.

The derived OpenCode comparison shows the same resource-efficiency pattern on the 50 experimental instances randomly sampled from SWE-bench Pro. Plain OpenCode resolves 30 of 50 instances (60.0\%), while ContextSniper resolves 29 of 50 (58.0\%), a difference of one instance or 2.0 percentage points. Across the 50 experimental instances, ContextSniper reduces total token consumption from 94,173,981 to 56,549,708 tokens (39.95\%) and average rounds per instance from 39.98 to 28.73 (28.13\%). Thus, ContextSniper retains a small observed correctness deficit while using approximately 37.62 million fewer tokens and substantially fewer interaction rounds. Because the comparison incorporates two accepted re-runs and provides one retained observation per instance, it does not establish statistical or causal superiority.

\begin{figure}[H]
\centering
\begin{tikzpicture}
\begin{axis}[
    width=0.78\textwidth,
    height=2.45in,
    title={Within-Host Efficiency Reduction},
    xlabel={Host agent},
    ylabel={Reduction vs. baseline (\%)},
    ymin=0,
    ymax=65,
    symbolic x coords={OpenClaw,OpenCode},
    xtick=data,
    ymajorgrids=true,
    grid style={csraw!25},
    bar width=22pt,
    title style={font=\bfseries},
    enlarge x limits=0.35,
    legend style={at={(0.5,-0.22)},anchor=north,legend columns=2,draw=none},
    legend cell align={left},
    area legend
]
\addplot[ybar, bar shift=-12pt, fill=cssniper, draw=cssniper!70!black]
table[x=host,y=token_reduction,col sep=comma] {resource/cross-host-efficiency.csv};
\addplot[ybar, bar shift=12pt, fill=csband, draw=csblue!70!black]
table[x=host,y=cost_reduction,col sep=comma] {resource/cross-host-efficiency.csv};
\node[font=\bfseries\small, yshift=5pt, xshift=-12pt] at (axis cs:OpenClaw,\OpenClawTokenReduction) {\OpenClawTokenReduction};
\node[font=\bfseries\small, yshift=5pt, xshift=12pt] at (axis cs:OpenClaw,\OpenClawCostReduction) {\OpenClawCostReduction};
\node[font=\bfseries\small, yshift=5pt, xshift=-12pt] at (axis cs:OpenCode,\OpenCodeTokenReduction) {\OpenCodeTokenReduction};
\node[font=\bfseries\small, yshift=5pt, xshift=12pt] at (axis cs:OpenCode,\OpenCodeCostReduction) {\OpenCodeCostReduction};
\legend{Token reduction, Cost reduction}
\end{axis}
\end{tikzpicture}
\caption{Within-host efficiency reduction relative to each host agent's baseline setting. Absolute token and cost values are reported in Table~\ref{tab:main-results}.}
\label{fig:cross-host-reduction}
\end{figure}
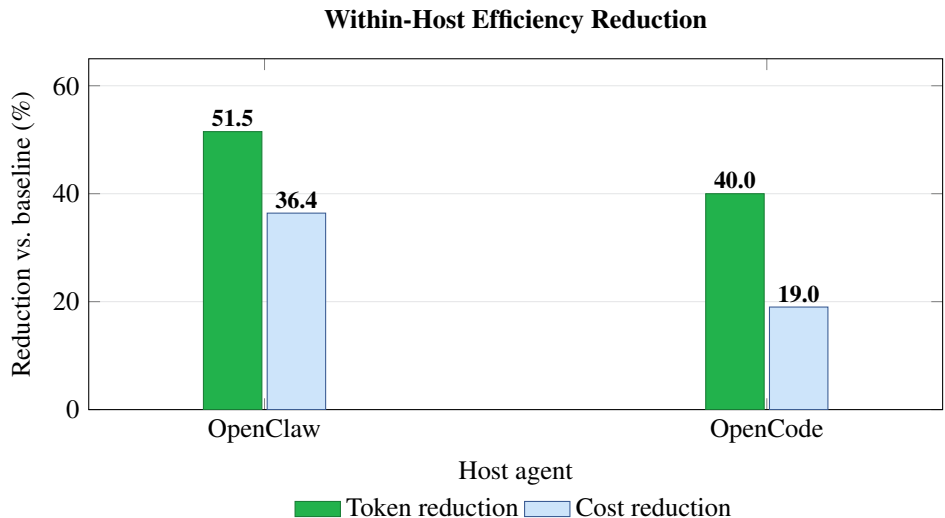

\begin{table}[tbp]
\centering
\footnotesize
\setlength{\tabcolsep}{1.0pt}
\caption{Main efficiency and validation results for OpenClaw on SWE-bench Lite and OpenCode on SWE-bench Pro. Each method completed 50 experimental runs. `Val. OK' reports validation runs that completed without validator error, not agent task completion. `Avg. Int.' is the host-native interaction measure: actions for OpenClaw and \texttt{step\_finish} rounds for OpenCode; values are comparable only within a host. `Ctx. Act.' counts context-acquisition actions and is available only for OpenClaw. Dashes denote metrics not reported by the corresponding experiment.}
\label{tab:main-results}
\begin{filecontents*}{resource/main-results.csv}
benchmark,host,method,runs,patch,val_ok,res,res_rate,total_tok,cost,avg_int,ctx_act,rel_tok
Lite,OpenClaw,Baseline,50,50,26,13,26.0\%,67.80M,\$0.635,32.1,572,--
Lite,OpenClaw,ContextSniper,50,50,33,12,24.0\%,32.91M,\$0.404,17.2,461,-51.5\%
Lite,OpenClaw,Rel. change,--,--,+7,-1,-2.0 pp,-51.5\%,-36.4\%,-46.4\%,-19.4\%,--
Pro,OpenCode,Baseline,50,50,50,30,60.0\%,94.17M,\$0.0147,39.98,--,--
Pro,OpenCode,ContextSniper,50,50,50,29,58.0\%,56.55M,\$0.0119,28.73,--,-40.0\%
Pro,OpenCode,Rel. change,--,--,0,-1,-2.0 pp,-40.0\%,-19.0\%,-28.1\%,--,--
\end{filecontents*}
\pgfplotstabletypeset[
  col sep=comma,
  string type,
  columns={benchmark,host,method,runs,patch,val_ok,res,res_rate,total_tok,cost,avg_int,ctx_act,rel_tok},
  columns/benchmark/.style={column name=Benchmark},
  columns/host/.style={column name=Host},
  columns/method/.style={column name=Method},
  columns/runs/.style={column name=Runs},
  columns/patch/.style={column name=Patch},
  columns/val_ok/.style={column name=Val. OK},
  columns/res/.style={column name=Res.},
  columns/res_rate/.style={column name=Res. Rate},
  columns/total_tok/.style={column name=Total Tok.},
  columns/cost/.style={column name=Cost},
  columns/avg_int/.style={column name=Avg. Int.},
  columns/ctx_act/.style={column name=Ctx. Act.},
  columns/rel_tok/.style={column name=Rel. Tok.},
  every head row/.style={before row=\hline, after row=\hline},
  every last row/.style={after row=\hline}
]{resource/main-results.csv}
\end{table}

Table~\ref{tab:main-results} reports the metrics shared by the two 50-instance experiments while preserving host-specific measurement definitions. The OpenCode experiment reports 50 submitted patches in both conditions and logged costs of \$0.0147 for the baseline and \$0.0119 for ContextSniper, a 19.0\% reduction. It does not provide a context-acquisition action count, so those cells remain unreported. A five-task Django pilot with additional retrieval and memory systems is useful as exploratory evidence, but it is not used as a main competitor claim because it does not yet match the 50-task protocol.

\begin{figure}[H]
\centering
\begin{tikzpicture}
\begin{axis}[
    width=0.82\textwidth,
    height=2.35in,
    title={Per-Task Token Usage},
    xlabel={Baseline total tokens (M)},
    ylabel={ContextSniper total tokens (M)},
    xmode=log,
    ymode=log,
    log basis x=10,
    log basis y=10,
    xmin=0.05,
    xmax=30,
    ymin=0.05,
    ymax=30,
    xtick={0.1,0.3,1,3,10,30},
    ytick={0.1,0.3,1,3,10,30},
    xticklabels={0.1,0.3,1,3,10,30},
    yticklabels={0.1,0.3,1,3,10,30},
    ymajorgrids=true,
    xmajorgrids=true,
    grid style={csraw!25},
    title style={font=\bfseries},
    legend style={at={(0.5,-0.24)},anchor=north,legend columns=3,draw=none},
    legend cell align={left},
    unbounded coords=discard
]
\addplot[only marks, mark=*, mark size=1.7pt, cssniper, opacity=0.72]
table[
    x=legacy_m,
    y=context_sniper_m,
    col sep=comma,
    restrict expr to domain={\thisrow{host_id}}{0:0}
] {resource/per-task-token-usage.csv};
\addplot[only marks, mark=square*, mark size=2.0pt, csblue, opacity=0.72]
table[
    x=legacy_m,
    y=context_sniper_m,
    col sep=comma,
    restrict expr to domain={\thisrow{host_id}}{2:2}
] {resource/per-task-token-usage.csv};
\addplot[csraw!70, dashed, thick] coordinates {(0.05,0.05) (30,30)};
\legend{OpenClaw (Lite), OpenCode (Pro), Equal token use}
\end{axis}
\end{tikzpicture}
\caption{Per-task token-usage comparison on matched benchmark tasks: 50 SWE-bench Lite tasks for OpenClaw and 41 SWE-bench Pro tasks for OpenCode. Each point compares the host-agent baseline and ContextSniper for one task. Both axes use a logarithmic scale; points below the diagonal indicate tasks where ContextSniper uses fewer tokens.}
\label{fig:per-task-token-usage}
\end{figure}
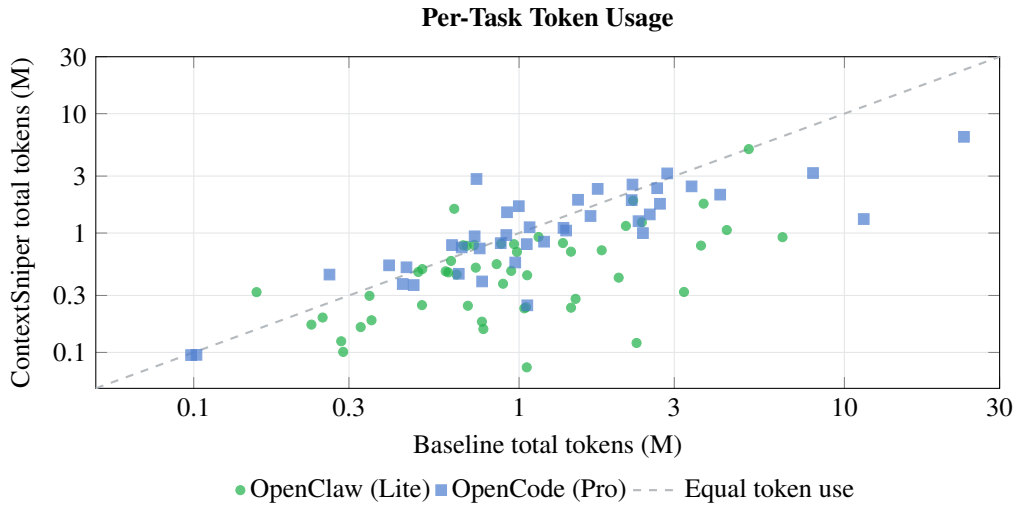

\begin{figure}[!t]
\centering
\begin{tikzpicture}
\begin{axis}[
    width=0.82\textwidth,
    height=2.12in,
    title={(a) OpenClaw on SWE-bench Lite},
    ylabel={Token change vs. baseline (\%)},
    ymin=-90,
    ymax=10,
    ytick={-80,-60,-40,-20,0},
    symbolic x coords={django,sphinx,scikit-learn,psf,sympy,mwaskom,pylint,pydata,astropy,pallets,matplotlib,pytest},
    xtick=data,
    x tick label style={rotate=32,anchor=east,font=\scriptsize},
    ymajorgrids=true,
    grid style={csraw!25},
    title style={font=\bfseries},
    enlarge x limits=0.05,
    unbounded coords=discard
]
\addplot[ybar, fill=cssniper, draw=cssniper!70!black, bar width=8pt]
table[
    x=repo,
    y=token_change,
    col sep=comma,
    restrict expr to domain={\thisrow{host_id}}{0:0}
] {resource/repo-token-change.csv};
\addplot[csraw!70, dashed, thick, forget plot] coordinates {(django,0) (pytest,0)};
\end{axis}
\end{tikzpicture}
\par\vspace{0.35em}
\begin{tikzpicture}
\begin{axis}[
    width=0.82\textwidth,
    height=2.12in,
    title={(b) OpenCode on SWE-bench Pro},
    xlabel={Repository},
    ylabel={Token change vs. baseline (\%)},
    ymin=-90,
    ymax=10,
    ytick={-80,-60,-40,-20,0},
    symbolic x coords={vuls,ansible,NodeBB,qutebrowser,openlibrary,teleport,element-web,navidrome,tutanota},
    xtick=data,
    x tick label style={rotate=32,anchor=east,font=\scriptsize},
    ymajorgrids=true,
    grid style={csraw!25},
    title style={font=\bfseries},
    enlarge x limits=0.06,
    unbounded coords=discard
]
\addplot[ybar, fill=csblue, draw=csblue!70!black, bar width=9pt]
table[
    x=repo,
    y=token_change,
    col sep=comma
] {resource/opencode-repo-token-change.csv};
\addplot[csraw!70, dashed, thick, forget plot] coordinates {(vuls,0) (tutanota,0)};
\end{axis}
\end{tikzpicture}
\caption{Repository-level token change on the matched benchmark task sets for (a) OpenClaw on SWE-bench Lite and (b) OpenCode on SWE-bench Pro. Values aggregate total tokens by repository before computing relative change against each host-agent baseline. Negative values indicate token savings from ContextSniper. Repositories in panel (b) are ordered from the largest token reduction to the positive token increase.}
\label{fig:repo-token-change}
\end{figure}
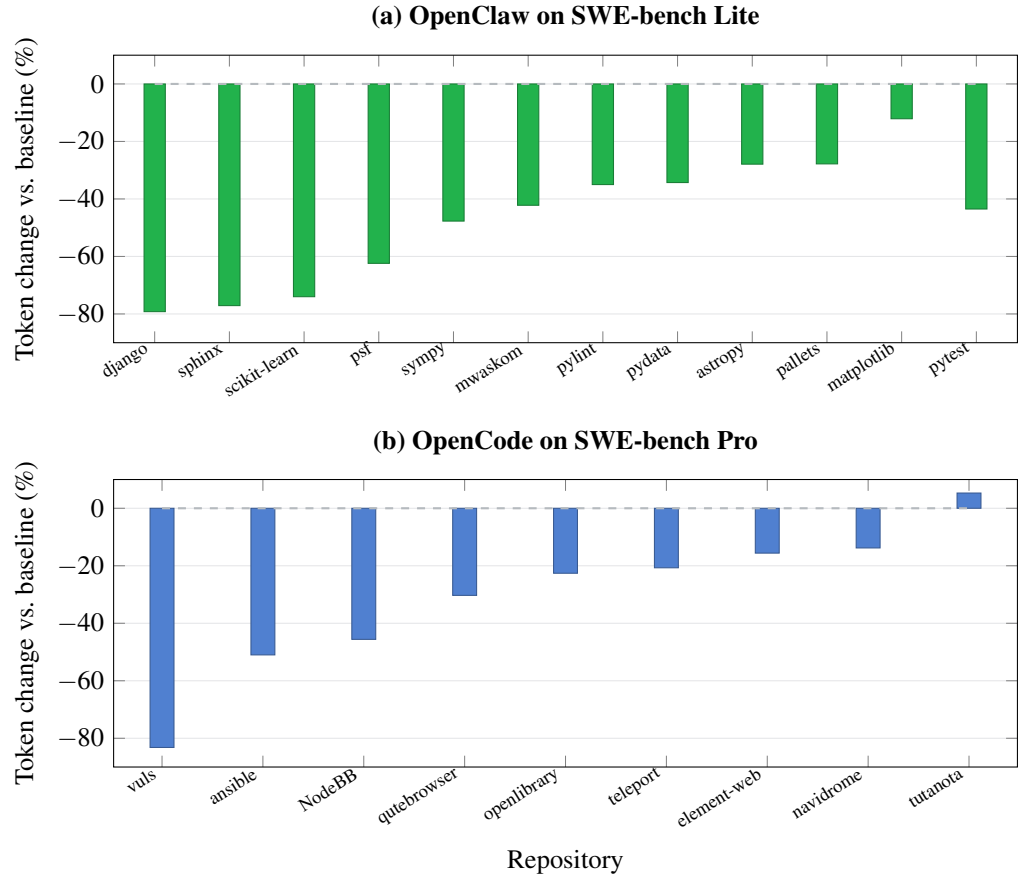

\subsection{Task-Level Failure Analysis}
The matched OpenClaw comparison shows the intended behavior on most tasks: ContextSniper uses fewer tokens on \resultcell{token_saving_tasks} of \resultcell{matched_tasks} matched tasks. The remaining \resultcell{token_increase_tasks} cases are more informative than the aggregate average because they show when a context-access layer can add overhead rather than remove it. Trace inspection separates these increases into two patterns.

First, some increases are dominated by fixed retrieval and tool-interface overhead. On \texttt{pylint-dev\_\_pylint-7228}, \texttt{pallets\_\_flask-5063}, and \texttt{matplotlib\_\_matplotlib-24265}, ContextSniper uses the same number of total actions as the OpenClaw baseline, but adds two to four code-search calls. The resulting token increases are modest, between 9.6\% and 17.3\%. These are not cases where the agent spirals into much longer exploration; rather, the task is short enough that the retrieval layer's setup and search turns are not fully amortized by shorter later context.

Second, larger increases occur when retrieval does not quickly end exploration or when the baseline run is unusually short. The clearest retrieval-overhead failure is \texttt{mwaskom\_\_seaborn-3407}: ContextSniper uses six code-search calls and 32 actions, compared with 21 baseline actions, raising token use by 152.3\%. This suggests a retrieval miss or insufficiently decisive evidence packet that caused the host agent to continue searching. In \texttt{pytest-dev\_\_pytest-7168}, ContextSniper retrieves the relevant \texttt{saferepr.py} region early, but the baseline completes the task in only five actions; even after rerun, the ContextSniper path uses nine actions and two search calls, making the relative token increase 104.6\%. This is an overhead-on-short-task case rather than a pure localization failure. Separately, \texttt{pytest-dev\_\_pytest-7220} was a large pre-rerun increase, but after rerun it becomes a token reduction; the accepted trace still records backend startup warnings, so we exclude it from the current token-increase set.

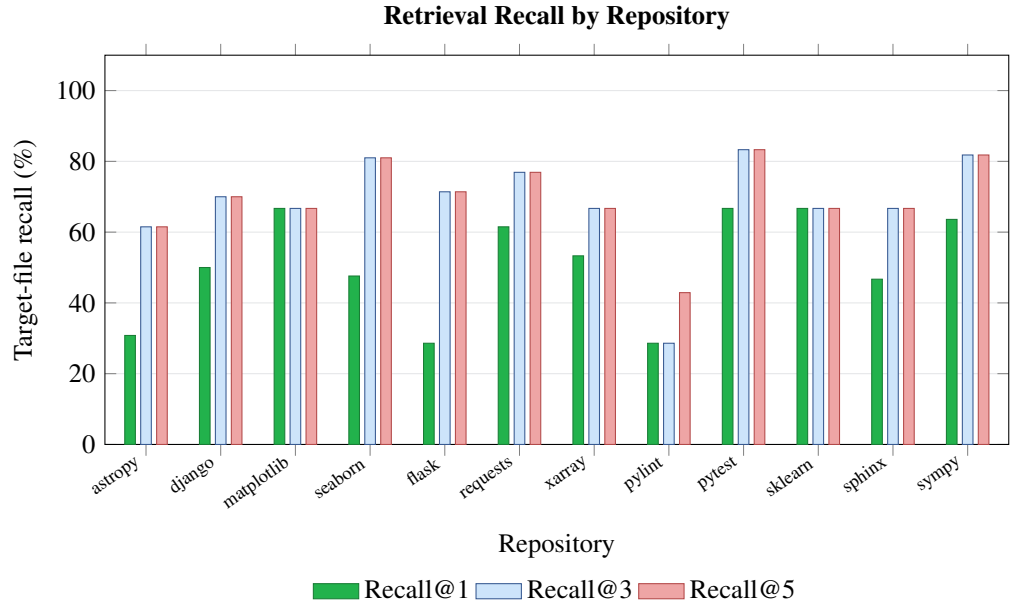
\begin{figure}[H]
\centering
\begin{tikzpicture}
\begin{axis}[
    width=0.82\textwidth,
    height=2.65in,
    title={Retrieval Recall by Repository},
    xlabel={Repository},
    ylabel={Target-file recall (\%)},
    ymin=0,
    ymax=110,
    symbolic x coords={astropy,django,matplotlib,seaborn,flask,requests,xarray,pylint,pytest,sklearn,sphinx,sympy},
    xtick=data,
    x tick label style={rotate=35,anchor=east,font=\scriptsize},
    ymajorgrids=true,
    grid style={csraw!25},
    ybar,
    bar width=4pt,
    title style={font=\bfseries},
    enlarge x limits=0.05,
    legend style={at={(0.5,-0.32)},anchor=north,legend columns=3,draw=none},
    legend cell align={left},
    area legend
]
\addplot[fill=cssniper, draw=cssniper!70!black]
table[x=repo,y=recall_at_1,col sep=comma] {resource/retrieval-recall.csv};
\addplot[fill=csband, draw=csblue!70!black]
table[x=repo,y=recall_at_3,col sep=comma] {resource/retrieval-recall.csv};
\addplot[fill=csmark!55, draw=csmark!80!black]
table[x=repo,y=recall_at_5,col sep=comma] {resource/retrieval-recall.csv};
\legend{Recall@1, Recall@3, Recall@5}
\end{axis}
\end{tikzpicture}
\caption{Retrieval recall by repository from actual ContextSniper search samples labeled with benchmark patch target files. The denominator is search attempts per repository; Recall@k reports whether a target file appears in the top-k returned paths.}
\label{fig:retrieval-recall-by-repo}
\end{figure}

Figure~\ref{fig:retrieval-recall-by-repo} complements this task-level analysis by measuring whether the retriever surfaces the eventual target file within the first returned paths. Low-recall repositories identify where the first failure pattern is more likely: if the target file is absent from the early results, the host agent must either issue more ContextSniper searches or fall back to broader native exploration. The token-increase cases therefore do not overturn the efficiency result, but they calibrate it: ContextSniper is strongest when early retrieval replaces broad exploration, and weakest when retrieval overhead is added to an already short baseline run or when the first evidence packet misses the decisive file.

\subsection{Comparison with Memory and RAG Systems}
To contextualize ContextSniper against related memory-style integrations, we run a five-task Django pilot using the same OpenClaw host setting. The pilot compares the OpenClaw baseline, ContextSniper, and several memory or retrieval integrations, including mem0, Letta/MemGPT, OpenViking, TencentDB Memory, Serena, and a LlamaIndex-style RAG baseline \parencite{mem02026github,letta2026github,openviking2026github,tencentdbmemory2026github,serena2026github,llamaindex2026github}. All methods complete five task runs and are evaluated with official SWE-bench validation reports.

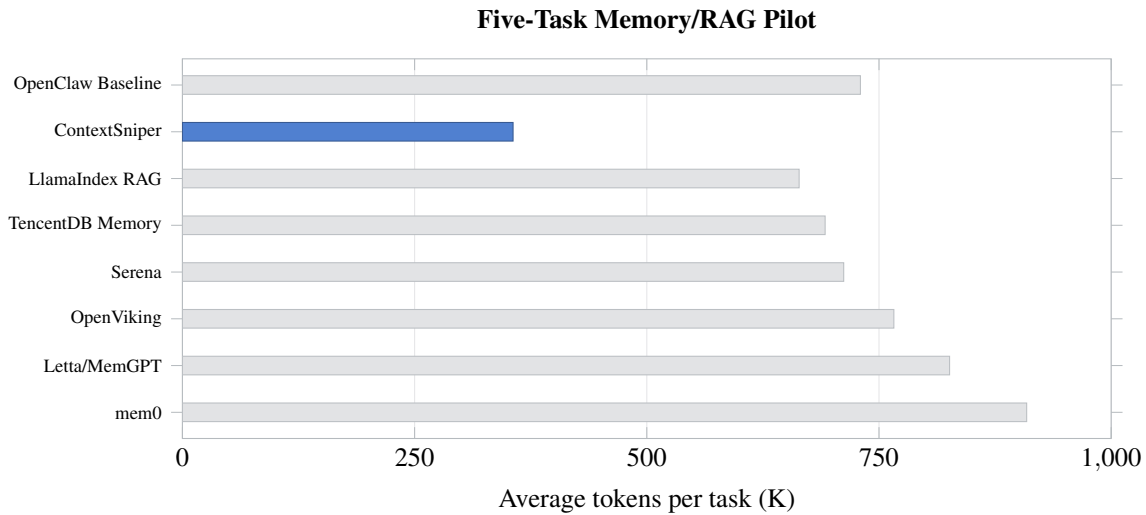
\begin{figure}[H]
\centering
\begin{tikzpicture}
\begin{axis}[
    xbar,
    width=0.84\textwidth,
    height=2.6in,
    title={Five-Task Memory/RAG Pilot},
    xlabel={Average tokens per task (K)},
    xmin=0,
    xmax=1000,
    xtick={0,250,500,750,1000},
    symbolic y coords={mem0,Letta/MemGPT,OpenViking,Serena,TencentDB Memory,LlamaIndex RAG,ContextSniper,OpenClaw Baseline},
    ytick={mem0,Letta/MemGPT,OpenViking,Serena,TencentDB Memory,LlamaIndex RAG,ContextSniper,OpenClaw Baseline},
    y tick label style={font=\scriptsize},
    xmajorgrids=true,
    grid style={csraw!25},
    axis line style={csraw!70},
    tick style={csraw!70},
    bar width=7pt,
    title style={font=\bfseries},
    enlarge y limits=0.08,
]
\addplot[xbar, bar shift=0pt, fill=csraw!28, draw=csraw!65]
table[
    x=avg_tokens_k,
    y=method_id,
    col sep=comma,
    restrict expr to domain={\coordindex}{0:0}
] {resource/memory-system-pilot.csv};
\addplot[xbar, bar shift=0pt, fill=csraw!28, draw=csraw!65]
table[
    x=avg_tokens_k,
    y=method_id,
    col sep=comma,
    restrict expr to domain={\coordindex}{2:7}
] {resource/memory-system-pilot.csv};
\addplot[xbar, bar shift=0pt, fill=csblue, draw=csblue!70!black]
table[
    x=avg_tokens_k,
    y=method_id,
    col sep=comma,
    restrict expr to domain={\coordindex}{1:1}
] {resource/memory-system-pilot.csv};
\end{axis}
\end{tikzpicture}
\caption{Five-task Django pilot comparison with memory and RAG-style integrations. Bar length shows average logged tokens per task. The pilot is exploratory and smaller than the main 50-task OpenClaw comparison; task-resolution counts are reported in the text.}
\label{fig:memory-system-pilot}
\end{figure}

Figure~\ref{fig:memory-system-pilot} should be read as exploratory evidence rather than the main benchmark because its five-task size is much smaller than the matched OpenClaw comparison in Table~\ref{tab:main-results}. Task outcomes are similar: all methods resolve four of five pilot tasks except OpenViking, which resolves five of five. We therefore use the figure to compare token-efficiency patterns, not to make a final solve-rate claim.

\section{Discussion}
\fontsize{10}{12}\selectfont

\subsection{ContextSniper as AntTrail Infrastructure}
ContextSniper treats code memory and context control as infrastructure for coding agents. Its central design choice is to retrieve broadly, expose narrowly, and preserve recoverable context outside the prompt. Placing this memory layer outside the host agent's reasoning loop lets the same substrate support different tool interfaces, as demonstrated by the OpenClaw pilot adapter.

This positioning also clarifies the role of the "sniper" metaphor. ContextSniper is not simply a larger index or a summarizer for entire repositories. Its job is to remove unwanted context before the LLM spends attention on it: irrelevant file regions, redundant exploration, noisy command output, stale snippets, and broad repository dumps. The system is useful when it keeps the concrete evidence that matters for a repair while preventing surrounding noise from filling the context window.

The result is a division of labor between agent and infrastructure. The host agent decides what problem to solve and what patch to write. ContextSniper decides how to expose repository evidence under a limited context budget. This makes context management an explicit systems layer rather than an implicit side effect of file reads and shell commands.

\subsection{Deployment}
ContextSniper can be deployed through host-agent plugins because its boundary is intentionally narrow: it mediates context access while leaving planning, editing, and validation with the host agent. The pilot adapters in this paper instantiate that boundary for two tool surfaces, but the released repository is best understood as the testing and evaluation harness around the specialization rather than as the full memory engine.

The AntTrail framing matters for scope. This paper evaluates the repository-level repair setting, while the parent memory engine targets broader cross-domain retrieval and context management. The results therefore characterize the coding specialization, not the complete AntTrail system.

\subsection{Practical Implications}
The most direct implication is lower token cost. If the agent receives fewer irrelevant file regions and shorter command outputs, it pays for less input context and often needs fewer exploratory turns. This matters for repository-level repair because the expensive part of many runs is not only patch generation, but the repeated process of finding the files, symbols, and tests that matter.

ContextSniper also makes the agent's working context more predictable. Raw file reads and shell commands can vary widely in size, and a single broad output can dominate a turn. Context control gives the system a place to enforce budgets and remove output that is unlikely to help. The same logs can make debugging easier: developers can inspect which snippets were retrieved, which outputs were suppressed, and whether the final patch touched files that ContextSniper surfaced early.

For local-codebase workflows, this design preserves the agent's normal interaction style. Developers do not need to manually curate a prompt for each issue, and the agent does not need to ingest an entire repository. Instead, ContextSniper provides a local memory layer that can be queried and refreshed as the repository changes. This is especially useful for private or evolving codebases where relevant evidence is spread across implementation files, tests, configuration, and runtime output.

\subsection{Limitations and Future Work}
Broader conclusions about repair success require more standardized validation, more tasks, and more repositories. A SWE-bench Verified evaluation would strengthen the evidence, and future work should report confidence intervals so token reductions can be separated from benchmark and validation noise. The present results should also not be read as an evaluation of the full cross-domain AntTrail engine.

ContextSniper also depends on retrieval and indexing quality. Embedding models, chunking strategies, symbol extraction, and structural links all affect which snippets are surfaced. If the relevant evidence is weakly signaled, spread across unusual files, or visible only after running tests, retrieval may miss it. Better hybrid retrieval, subtask-aware recall, and failure-aware reranking are natural extensions.

Context control introduces its own risk. The system can remove useful evidence if the sniping policy is too aggressive or if the task requires broad architectural context. Agents may need an explicit way to request broader output, expand a shortened region, or mark a file as important. Future work should study learned sniping policies, budget sweeps, and human-inspectable traces that explain why particular snippets were kept or removed. Commit-history memory is another promising direction because prior changes and linked issues may reveal repair-relevant files that current-source retrieval alone misses.

\section{Conclusion}
\fontsize{10}{12}\selectfont

ContextSniper makes repository code-context access more token-efficient by removing unwanted context and preserving repair-relevant evidence. The disclosed OpenClaw results show substantial reductions in context cost while keeping the host agent's normal workflow. More broadly, the paper argues that coding agents should treat code memory and context control as explicit infrastructure, not as incidental byproducts of raw file reads and shell commands.

\vspace{0.5cm}

\balance
\bibliographystyle{IEEEtran}
\bibliography{references}

\end{document}